\documentclass[sn-mathphys,Numbered]{sn-jnl}

\usepackage{graphicx}%
\usepackage{multirow}%
\usepackage{amsmath,amssymb,amsfonts}%
\usepackage{amsthm}%
\usepackage{mathrsfs}%
\usepackage[title]{appendix}%
\usepackage{xcolor}%
\usepackage{textcomp}%
\usepackage{manyfoot}%
\usepackage{booktabs}%
\usepackage{algorithm}%
\usepackage{algorithmicx}%
\usepackage{algpseudocode}%
\usepackage{listings}
\usepackage{url}
\usepackage{comment}
\usepackage{natbib}
\setcitestyle{authoryear,open={(},close={)}} 
\newtheorem{hypothesis}{Hypothesis}

\newtheorem{remark}{Remark}%

\begin{document}

This work has been submitted to User Modeling and User-Adapted Interaction by Springer for possible publication. Copyright may be transferred without notice, after which this version may no longer be accessible.

\newpage

\title[Culture-to-Culture Image Translation and User Evaluation]{Culture-to-Culture Image Translation and User Evaluation}


\author[1]{\fnm{Giulia Zaino} \sur{Author}}\email{giuliazaino46@gmail.com}
\equalcont{Authors contributed equally to this work.}

\author[1]{\fnm{Carmine Recchiuto} \sur{Author}}\email{carmine.recchiuto@dibris.unige.it}
\equalcont{Authors contributed equally to this work.}

\author*[1]{\fnm{Antonio Sgorbissa} \sur{Author}}\email{antonio.sgorbissa@unige.it}
\equalcont{Authors contributed equally to this work.}

\affil*[1]{\orgdiv{DIBRIS}, \orgname{University of Genova}, \orgaddress{\street{Via Opera Pia 13}, \city{Genova}, \postcode{16145}, \country{Italy}}}


\abstract{The article introduces the concept of image ``culturization," which we define as the process of altering the ``brushstroke of cultural features" that make objects perceived as belonging to a given culture while preserving their functionalities. First, we defined a pipeline for translating objects' images from a source to a target cultural domain based on state-of-the-art Generative Adversarial Networks. Then, we gathered data through an online questionnaire to test four hypotheses concerning the impact of images belonging to different cultural domains on Italian participants. 
As expected, results depend on individual tastes and preferences: however, they align with our conjecture that some people, during the interaction with an intelligent system, will prefer to be shown images modified to match their cultural background.
The study has two main limitations. First, we focussed on the culturization of individual objects instead of complete scenes. However, objects play a crucial role in conveying cultural meanings and can strongly influence how an image is perceived within a specific cultural context. Understanding and addressing object-level translation is a vital step toward achieving more comprehensive scene-level translation in future research. Second, we performed experiments with Italian participants only. We think that there are unique aspects of Italian culture that make it an interesting and relevant case study for exploring the impact of image culturization. Italy is a very culturally conservative society, 
and Italians have specific sensitivities and expectations regarding the accurate representation of their cultural identity and traditions, which can shape individuals' preferences and inclinations toward certain visual styles, aesthetics, and design choices. As a consequence, we think they are an ideal candidate for a preliminary investigation of how image culturization affects participants' responses. 

}




\maketitle

\section{Introduction}
\label{Introduction}
{Y}{ukiko} is an 83-year-old Japanese woman living with her son Matsuo in a traditional Japanese house. 
Yukiko sleeps on a futon on the tatami floor, which, she says, is very good for her back. Recently, every time Yukiko has woken up at night to go to the bathroom, she has felt a little dizzy and confused, and sometimes she has had trouble finding the light switch. 
Last week she fell in the dark, and nobody noticed her lying on the floor until morning. Matsuo is worried about her safety and decided to set up a Smart Environment composed of a small table-top robot assistant with a tablet-like display on its chest, voice-activated lights, and cameras that can detect emergencies. Yukiko gladly agreed to have a camera installed in her room: ``I don’t want to be a burden for my son," she says. 

It’s Sunday evening: a cartoon-like character on the tablet with the appearance of Tetsuwan Atomu\footnote{The android boy popular in Japan, also known as ``Astro Boy".} greets her with a bow, chats with her about cherry blossoms in Spring by showing pictures of trees in Kyoto, and then brushes his teeth with a colored toothbrush under the light of a Japanese paper lamp, suggesting to go to sleep. ``You know so many things," Yukiko says with a smile after brushing her teeth and before lying down on her futon on the tatami floor, ``Goodnight!" 

Tetsuwan receives an alert from one of the cameras recognizing a person lying on the floor: this could be an emergency, which ordinarily requires alerting Yukiko’s family. However, according to the robot's cultural database, sleeping on a tatami floor in a traditional Japanese house is common. ``Switch off the light, please," says Yukiko, yawning and confirming Tetsuwan’s assessment. Tetsuwan relaxes: ``Goodnight!", he replies.

The scenario above introduces a key concept: 
how to provide personalized interaction by taking the cultural context into account. Indeed, the way Tetsuwan Atomu greets Yukiko, the pictures it shows on its screen, and how it interprets the situation when Yukiko lies on the floor show that the assistant is culturally competent. 
The concept of ``culture" is complex \citep{Geertz73, Hofstede80, Leininger1988, Schwartz19921, Henare2007} 
 and there is no consensus among researchers in defining it. 
A simple yet effective definition holds that culture is a shared representation of the world of a group of people. 
Then, by ``culturally competent" \citep{Betancourt2003}, we mean an intelligent system that can adapt its visual representations, perceptions, plans, actions, and interaction style depending on the worldview of the person it is interacting with 
\citep{
Bruno2017, 
Bruno2019}. 

\begin{figure}
    \centering
    \includegraphics[width=0.81\columnwidth]{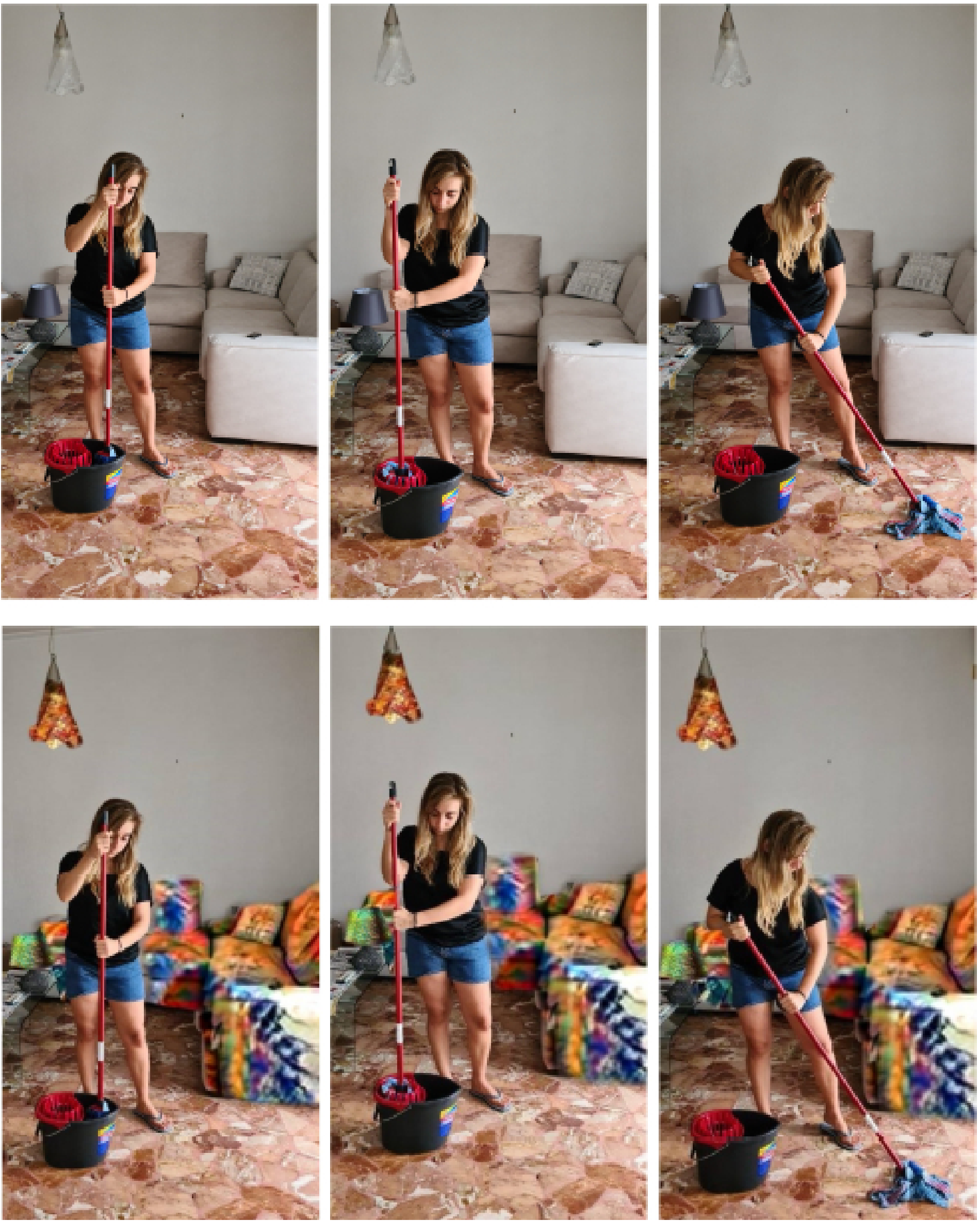}
    \caption{Instructions for dipping the mop, wringing it, and mopping with different ``cultural brushstrokes."}
    \label{fig:cleaning}
\end{figure}

In this general scenario, this article focuses on the visual representation of objects 
in different cultures \citep{Knappett2005} and the impact on people. The anthropological study of material culture teaches us that everyday objects have different designs in diverse world areas, even when they share many similarities in their functionalities. This is not a surprise: objects' design relates to the social tasks they intend to accomplish and the material enabling them to do it  \citep{Robb2015}. 
Consequently, within a given geographical area and culture, some visual features tend to coherently repeat with a higher frequency, even if aesthetic universals have been found \citep{Hekkert2008}. 
With a metaphor, two objects may have the same functionalities or affordances 
and yet be given ``brushstrokes of cultural features" that make them uniquely recognizable. 

Suppose now a robot 
showing instructions on performing a given task such as brushing one's teeth or cleaning the floor: this capability may play a key role in SARs, e.g., when interacting with older people with mild forms of dementia. 
In the case of floor cleaning, instructions are likely to go like this: (1) Take a bucket; (2) Take a mop; (3) Take a cleaning detergent; (4) Sweep or vacuum First; (5) Fill the bucket; (6)  Dip the mop; (7) Wring the mop; (8) Begin mopping (steps 6, 7, and 8 are shown in Figure \ref{fig:cleaning}). 
How important is it that, during interaction with people, 
robots and other intelligent systems show images with which the person is familiar (such as by adding ``Moroccan brushstrokes" in the bottom images in Figure \ref{fig:cleaning})? The advertising industry knows well that a product may need culturally appropriate advertisements to hit the market in diverse world areas. 
An ad for a detergent in 
Japan is unlikely to 
show an Italian family in an Italian house unless a specific message in this sense is required, and the opposite is true in Italy. This concept is the basis of cross-cultural design \citep{Hsu2011, Chai2015}, a process that requires designers to understand other cultures, select cultural elements, rethink and review them, and then integrate these into a new product to satisfy target users and incorporate them into product design.

Given these premises, the article's contribution is twofold.

\begin{itemize}
\item The article introduces the problem of ``culturizing" objects images and explores Generative Adversarial Networks (GAN) \citep{Goodfellow2014, Isola2017, Jun-Yan2017, Taeksoo2017, Mejjati2018, Sangwoo2018, Unit2017} for this purpose. 
With the term ``culturization," we intuitively mean translating selected objects in the image from a source to a target cultural domain, i.e., altering the brushstroke of cultural features that make objects perceived as belonging to a given culture while preserving their functionalities or affordances. 
\item The article explores the research question: `to what extent do people like images that were culturized to make them coherent with their culture?' To this end, we prepared an online questionnaire to be submitted to Italian participants. Then, we evaluated four hypotheses through statistical analysis: if Italian participants correctly recognize European objects (Hypothesis 1); if they prefer European objects (Hypothesis 2); if participants who prefer European objects also prefer non-European objects to be modified to look European (Hypothesis 3); if they prefer pictures of domestic environments that have been modified by culturizing pre-existing objects rather than  superimposing new objects downloaded from the internet on the picture (Hypothesis 4).

\end{itemize}

Remark that this work's novelty does not consist in proposing a new solution for image-to-image translation but in introducing the culturization concept, choosing a state-of-the-art GAN and a pipeline for image culturization, and implementing an experimental protocol for the subjective evaluation of culturized images with recruited participants.

The study has two main limitations. First, we focussed on the culturization of individual objects instead of complete scenes. However, objects play a crucial role in conveying cultural meanings and can strongly influence how an image is perceived within a specific cultural context. Understanding and addressing object-level translation is a vital step toward achieving more comprehensive scene-level translation in future research. Second, we performed experiments with Italian participants only. We think that there are unique aspects of Italian culture that make it an interesting and relevant case study for exploring the impact of image culturization. According to Schwartz's value framework \citep{Schwartz19921}, Italy is a culturally conservative society, with a lesser openness to novelty - for example concerning the intention to buy innovative products \citep{Rubera2011459}.  
Italians tend to have specific sensitivities and expectations regarding the accurate representation of their cultural identity and traditions, which can shape individuals' preferences and inclinations toward certain visual styles, aesthetics, and design choices. As a consequence, we think they are an ideal candidate for a preliminary investigation of how image culturization affects participants' responses. However, since the principles underpinning this research are general, they could potentially apply to different cultural contexts, ensuring the potential transferability of findings to other cultures through future research to validate and extend them across diverse cultural groups.

The rest of the article is organized as follows.
Section  \ref{sec:SOTA} describes the state-of-the-art related to cross-cultural design and image-to-image translation using GANs. Section \ref{Materials and Methods} presents the process of culturizing objects and environments and evaluating people's preferences. Sections \ref{results} and \ref{Discussion} present and discuss results. Conclusions are in Section \ref{sec:conclusion}.

\section{Background}
\label{sec:SOTA}

\subsection{Cross-cultural design}
\label{sec:Cross-cultural design}

Cross-cultural design is the problem of applying culture to product design, typically by rethinking and reviewing the cultural elements and integrating them into a product. The last 15 years have seen an increasing number of publications in the field and dedicated conferences \citep{CCD2022}, especially related to the cross-cultural design of technology products \citep{Lin2007146}. 
The survey in \citep{Chavan200926} analyzed mistakes designers make in preparing products for emerging markets by reporting examples from Kellog's and Whirpool's failed attempts to sell cereals or washing machines in India, and a similar analysis is performed in \citep{Winschiers-Theophilus2009665}. Along the same line, \citep{PatrickRau20121} argued that IT products and services are often produced and marketed across geographical 
boundaries without adequate consideration of culture, with a high probability that those developed in one country may not be effectively used in another country. Analyzing mobile text messaging use in American and Chinese contexts, \citep{Sun20121} observed that a technology created for a culturally localized user experience should mediate both instrumental practices and social meanings, starting from a 
rich understandings of the product's use in different contexts. 

According to these general principles, cross-cultural design has been explored in several areas. For example, the work in \citep{Lightner2002373} explored how e-commerce should take cultural factors into account by investigating the online shopping and behaviour preferences of Turkish and US students, 
and similar considerations, based on Hofstede's cultural dimensions \citep{Hofstede80}, were done with three major ethnicities in Indonesia \citep{Santoso2018}. 
Cross-cultural design in architecture has been addressed in \citep{Bonenberg2016105}, 
and the importance of a cross-cultural approach was highlighted in \citep{ORourke202289} by examining the effect of the physical environment in public hospitals and clinics on people’s perceptions and experiences of waiting for care. 
The work discussed in \citep{Pan2020308} explored traditional Chinese painting aesthetics to identify elements that can play a role in the modern cross-cultural design. 
Similar ideas were applied in \citep{Zhou2021373} to analyze the artistic and cultural characteristics of ancient Chinese ships as an inspiration for modern yacht design 
whereas \citep{Asino2019395} explored strategies to guide learning practitioners to understand the impact of culture in the design of learning materials. 
The study in \citep{Cao2021197} proposed design heuristics for sustainable packaging generation 
adapted to global users. 

To provide general theoretic frameworks and guidelines for cross-cultural design, \citep{Li2020779} measured the empathy 
of designers trying to understand users' experience from different cultural backgrounds, and 
\citep{Wang2020233} suggested that using in-situ making interlaced with evaluation is a feasible approach to drive designers 
in the early stage of design exploration. Based on cultural “onion and iceberg” models and the dimensions of human factors, \citep{Guo202220} proposed the “onion model” of human factors as a framework to guide cross-cultural design. 

\subsubsection{User Interfaces}
\label{sec:Interfaces}
User Interfaces are one of the areas where the impact of cultural factors has been explored the most. 
Just to mention a few examples, the study in \citep{Meier2014211}  explored how people from 18 different countries would gesture to control consumer electronics (e.g., TVs) 
and \citep{Jane20176794} compared how China vs. US users would intuitively interact with drones using gestures. 
The study in \citep{Urakami2019615} explored cultural differences in preferences for interface layouts between Japanese and Europeans concerning 
buttons and their positions. 
Regional differences in In-vehicle Information Systems design needs and preferences were explored across drivers from Australia and China in \citep{Young2012564}, whereas the cross-cultural design of consumer vehicles to improve safety is discussed in \citep{Koratpallikar2021539}. 
To generalize these concepts, the work in \citep{Heimg2021} proposed a toolbox for Intercultural User Interface Design and a methodology to create the link between 
Hofstede’s cultural dimensions and HCI dimensions (i.e., information density, information frequency, interaction speed, and frequency). In the same spirit, the CIAUI Framework \citep{Miraz2021199} specifically incorporated the concepts of universal design 
through AI-based adaptive interface development. 
The study in \citep{Miraz2022} presented a culturally inclusive 
mobile platform that takes snapshots of the installed apps on a smartphone as input, predicts the user’s cultural background, and offers a culturally customized user interface.

In \citep{SINGH2004864} 
websites from different countries were analyzed using Hofstede’s dimensions to propose policies for developing culturally congruent websites, whereas a cross-cultural analysis of Flickr users from Peru, Israel, Iran, Taiwan, and the UK was proposed in \citep{Dotan2010284}. The aim of \citep{
Alexander2017} was to propose strategies and guidelines for a cross-cultural website design 
and \citep{McMullen201619}  proposed a new approach for graphic designers working across cultural boundaries named Intercultural Design Competence. 
Finally, a literature review was proposed in \citep{Li2022105} focusing on 17 studies that compare design elements of Chinese and Western websites by finding differences in items and active elements, colors, links, and images. 



\subsubsection{Intelligent agents}
\label{sec:Robotics}

Cultural factors play a pivotal role in the interaction with intelligent agents and have been  studied in the field of human-robot interaction. 
The seminal work in \citep{Bartneck20051} 
presented a cross-cultural study on peoples' negative attitudes toward robots by recruiting 
participants from seven different countries.
The work in \citep{Shibata2009443} reported on the subjective evaluation of a seal robot 
in Japan, the UK, Sweden, Italy, South Korea, Brunei, and the US, whereas  \citep{Rau2010175} conducted a laboratory experiment 
to investigate the effects of culture, robot appearance, and task on human-robot interaction. 
The study in \citep{Mavridis2012517} explored attitudes toward robots in the Middle East 
using an Arabic-language conversational robot interacting with 
people from 38 countries in Dubai. 
A cross-cultural comparison of robot acceptance with Japanese and UK participants was performed in \citep{Nomura2015} and with Japanese and Australian participants in \citep{Haring2014166}. 
The research presented in \citep{Trovato2018} involved 
Peruvian and Japanese participants to investigate the perception of gender in robot design by manipulating body proportions and  \citep{Berque2022391} 
explored the role that kawaii (Japanese cuteness) plays in fostering acceptance of robots across cultures. 
The relations between robot anthropomorphism and cultural tendencies were also explored in \citep{
Dang2023} whereas \citep{Bernsteiner2022517} presented a cross-cultural comparative study 
with participants from Austria, Azerbaijan, Germany, and India on the perception of a humanoid, an animal-like, and a wheeled social robot. 


A very important research area explores the relations between culturally appropriate behaviour and the perceived trustworthiness of agents.
The authors of \citep{
Wang2010359} sought to clarify the effects of users' cultural background 
in accepting the choices made by a robot or a human assistant and \citep{Shidujaman2020, Bliss2020493} investigated how cultural familiarity affects acceptance of an agent's recommendation. 
In \citep{Andrist2015157}, two studies were presented comparing 
the credibility of robot speech between Arabic-speaking robots in Lebanon and English-speaking robots in the USA. The work in \citep{Rudovic2017} explored 
how behavioral engagement can vary across cultures in child-robot interaction for autism therapy  with Japanese and Serbian children whereas \citep{Makenova2018185} explored how persuasiveness in human-robot interaction can be influenced by 
robots' appearance and behavior, by conducting experiments 
with Kazakhstani citizens and foreign participants from Asia, Europe, and North America. 
Recently, we defined\footnote{https://cordis.europa.eu/article/id/124441-the-worlds-first-culturally-sensitive-robots-for-elderly-care}
a conceptual framework to make Socially Assistive Robots for elderly care culturally competent \citep{Bruno2017, Bruno2019} starting from research in Transcultural Nursing \citep{Leininger1988} and culturally competent Health Care \citep{Betancourt2003}.

\subsubsection{Visual elements and style-transfer}
\label{sec:Websites}

More closely related to the present work, cross-cultural aspects are crucial in visual representations, as shown in \citep{Knight200917}, which examined 
interpretations of icons and images from US websites with participants from Morocco, Sri Lanka, Turkey, and the USA. 
The contemporary printed graphic was investigated in \citep{McMullen20191}, revealing significant differences between countries in their use of layout, color, type, and images. 


When images need to be adapted to different cultural domains, style transfer can be a valuable method for combining two images into one, using the style of one image and the content of another. Interestingly, 
most of the following research has been done in China, confirming that the presence of cultural elements vs. novelty may play different roles in different cultures \citep{Rubera2011459}. For example, the authors of \citep{Quan2018} 
proposed to combine deep learning and Kansei engineering (a product development methodology that translates users' feelings, impressions, and emotions into concrete design parameters) to transfer a style image to a product's shape. A similar approach was proposed 
to produce cultural and creative products using different Chinese painting styles \citep{Yanlong2021721}, Nanjun Brocade \citep{Xuelin2021339}, and creative patterns on Indonesian Batik \citep{Joseph2021}. Interior decoration art was explored in \citep{Liu2021} and tools to design clothes with Dunhuang patterns and styles in \citep{Wu2021341}, 
whereas \citep{
Zhang2023} produced New Year Chinese prints in a new Pop art style and 
\citep{Fu2022670} 
painted objects with buildings having the distinctive features of Shanghai-style watercolor paintings. 
Finally, a tool is proposed in \citep{Zhou2022445} for designers to select and integrate cultural elements in the cross-cultural design process using deep learning techniques. The tool chooses the most suitable style image from a set of candidates and then applies the deep-learning-based style transfer technique to automatically produce a design image with the desired content and the cultural style of the selected style image.

\subsection{Image-to-image translation}

Despite the interesting results in cross-cultural design, style transfer aims to change the pictorial style of images 
rather than altering all the visual characteristics of an object (e.g., color, material, pattern, and form) to transform it into another object (e.g., a greek vase transformed into a Chinese vase or a Chinese lamp transformed into a Moroccan lamp).




The concept of image culturization is, therefore, different from style transfer and should rather be described as a more general image-to-image translation problem. 
Suppose having:

\begin{itemize}
\item a set of images related to each other by a shared characteristic that defines them as belonging to the same domain referred to as the source domain $S$;
\item a set of images, different from the first one, which defines a target domain $T$. 

\end{itemize}

The image-to-image translation problem is the problem of learning a mapping $G:S \to T$ such that the distribution of images from $G(S)$ is indistinguishable from the distribution $T$. Otherwise said, the objective is to modify images belonging to $S$ to make them ``similar" to those belonging to $T$ (in our case, by altering objects' color, material, pattern, and form). 

Image-to-image translation \citep{Kaji2019} may benefit from GAN-based approaches. 
GANs are not the only feasible approach to this problem \citep{pang2021imagetoimage}: however, we do not aim to find the most performing solution for image culturization. Instead, our objective is to find a feasible solution that meets all constraints, 
to be later tested with human participants to confirm the relevance of this new concept. According to this rationale, and considering that GAN-based methods are the vast majority of solutions, we will limit our analysis to GANs. 

 

\subsubsection{Generative Adversarial Networks}
\label{sec:GANstateoftheart}

GANs have been successful in several image-to-image translation domains, ranging from  the creation of purely synthetic images (e.g., faces of people that do not exist in the real world \citep{Goodfellow2014, radford2015unsupervised}) to the modification of selected features in a pre-existing image (e.g., the age or facial attributes of a person \citep{
Huang2021} 
or style-transfer (e.g., photorealistic images from sketched drawings \citep{Eitz2012} 
or paintings from photographs \citep{Jun-Yan2017}). Other GANs applications include adding semantic labels to photos 
\citep{Cordts2016} or filling missing regions of images 
\citep{Haofeng2019}. 

GANs have two components \citep{Goodfellow2014}: a generator and a discriminator. Both the generator and the discriminator must be trained using a large set of data $x$ describing a probability distribution $p_{data}$: the generator is trained to generate samples with probability distribution $p_g = p_{data}$; the discriminator is trained to distinguish the real data from the generated ones. The idea is to create a competition between two networks. 

In \citep{Goodfellow2014}, the discriminator and the generator are both multilayer perceptrons 
with parameters $\theta_g$ (the weights of the generator) and $\theta_d$ (the weights of the discriminator). 
The generator $x=G(z; \theta_g)$ takes as input a noise variable $z$ with probability distribution $p_z(z)$ and maps it to data space with a distribution $p_g(x )$. 
The discriminator is a binary classifier $D(x; \theta_d)$ that takes as input a sample $x$ (that may either be from one of the samples of the data or the output of the generator) and outputs the probability that $x$ came from the data distribution $p_{data}(x)$ rather than from the generator. After training, both $G$ and $D$ will reach a point at which they cannot improve because $p_{data}(x)=p_{g}(x)$, and  $D(x; \theta_d)=0.5$.
Optimal parameters $(\theta_g, \theta_d)$ are obtained by playing the following two-player minimax game with value function $V$, maximizing the cost function \eqref{eq:ganlossoriginal} over $\theta_d$ and minimizing it over $\theta_g$: 

\begin{equation}
\min_{\theta_g} \max_{\theta_d} V(\theta_d, \theta_g) = \frac{1}{2}\mathbb{E}_{x\sim p_{data}(x)}[logD(x)]+
 \frac{1}{2}\mathbb{E}_{z\sim p(z)}[1 - log(D(G(z)))]
 \label{eq:ganlossoriginal}
\end{equation}


Starting from this idea, Deep Convolutional Generative Adversarial Networks (DCGANs) were developed to exploit the success of Convolutional Neural Networks (CNNs)  \citep{radford2015unsupervised}: the generator still takes in input a noise variable $z$ but now generates an image through convolutional decoding operations. 
Unfortunately, similar to the original GANs, DCGANs do not consent to specify additional constraints on the generated data samples: a relationship between the noise  $z$ fed to the generator and the generated images exists (e.g., the network produces different faces) but we cannot control the outcome (e.g., we cannot choose to produce younger or older people's faces).
Conditional GANs (CGAN)  \citep{mirza2014conditional} overcome this limitation by providing additional information to the generator and the discriminator during training to synthesize a fake sample with desired characteristics. The generator learns to produce realistic samples (e.g., faces) corresponding to a specific label (e.g., younger or older), whereas the discriminator learns to distinguish fake sample-label pairs from real sample-label pairs. 
The value function $V$ is the same as Eq. \ref{eq:ganlossoriginal}, by substituting $G(z)$ with $G(z|y)$ and $D(x)$ with $D(x|y)$, where $y$ is the additional information provided to the network.
Conditional GANs can be the basis for image-to-image translation (end hence, image culturization).

In the following, we will review a subset of GAN-based solutions for image-to-image translation \citep{Isola2017, Jun-Yan2017, Taeksoo2017,  Mejjati2018, Sangwoo2018, Unit2017} that have interesting properties for image culturization. The analysis is limited to a small subset of the available solutions \citep{pang2021imagetoimage} since this article's main objective is to evaluate the impact of image culturization on people rather than comparing the performance of different GANs. However, given the general idea, we might consider other solutions in the future.

\begin{remark}
\label{R1}
We search for solutions capable of culturizing all design elements. 
There is a significant difference between color, material, and pattern on the one side and form on the other: changing the form of an object requires changing 
the subset of pixels belonging to the object and the background, which is known to pose significant challenges in image-to-image translation.  Some other elements may involve, for example, different tactile, olfactory, and auditory aspects, but they are all mapped to visual representations (or ignored) when focussing on images.
\end{remark}

\subsubsection{Pix2Pix}
\label{sec:Pix2Pix}
All approaches require two datasets, each containing images with homogeneous content, style, and resolution. However, some methods have the additional requirements that the images need to be paired during training: for every image of the source domain $S$, there should be a corresponding image of the target domain $T$. 
An example in this class is Pix2Pix \citep{Isola2017}. The model postulates a transformation that modifies the source image to make it ideally belong to the target domain while preserving some of its characteristics: the goal of Pix2Pix is to learn this transformation and perform it.
During training, Pix2Pix adopts a CGAN approach where the input image conditions the generator; the discriminator takes as input a generated or a real image belonging to the target domain and guesses whether the image is real or not. 
The network is trained with the CGAN loss function by adding a weighted term that measures the difference between the generated and real images in the pair.
The main limitation is that Pix2Pix and similar approaches require large paired datasets to learn the mapping between $S$ and $T$: paired datasets are complex and expensive to build, especially if one aims to culturize objects into several different cultures. %

\subsubsection{CycleGAN}
\label{sec:CycleGAN}
To overcome the constraint that datasets need to be paired during training, some approaches propose a new concept, referred to as \textit{consistency of the cycle}: the image $x_S$, after the first translation $G(x_S)$ from $S$ to $T$, is fed to a second generator that performs a backward translation $F(G(x_S))$ to remap the image to the starting domain $S$ and compares it with the original image.
This concept was first introduced by CycleGAN \citep{Jun-Yan2017}, 
trained by 
adding a weighed, ``cycle-consistency" L1 loss measuring the difference between $x_S$ and $F(G(x_S))$. 
CycleGAN and similar approaches \citep{Zili2017, Taeksoo2017} allow for building datasets quickly by searching for images on the web (e.g., Greek and Japanese vases). 
However, this solution also has negative sides. First, it cannot culturize an individual object in a complex scene and keep the background unaltered. Second, the image-to-image translation works better when changes occur in the color, material, and pattern (e.g., from a horse to a zebra) than when they occur in the form: CycleGAN can alter the form of an object, but this creates artifacts and less realistic images.

\subsubsection{Attention-guided GAN}
\label{sec:Attention-guided GAN}
Unlike the previous approaches, the solution proposed in attention-guided GAN \citep {Mejjati2018} learns which parts of the scene to translate from $S$ to $T$ by isolating a region of interest in the image. 
Based on the structure of CycleGAN and the concept of the consistency of the cycle, 
two \textit{attention networks} are added, one for the generator $G: S \to T$ and the other for the generator $F: T \to S$, which learn to extract ``attention maps." Attention maps are trained in parallel with generators and provide semantic information by segmenting images into regions that must or must not be translated from $S$ to $T$. This is achieved by assigning each pixel of the map a continuous value in the interval [0;1]: 
CycleGAN is equivalent to an attention-guided GAN where the attention maps equal $1$ everywhere.
This approach and similar ones \citep{Chen2018, Tang2021} 
allow modifying an individual scene element without further constraints on the datasets. However, 
attention maps imply a higher computational load and training time. Attention-guided approaches have limitations when image-to-image translation involves changes in the objects' form (typically required in image culturization, see the vases on the first row of Figure \ref{all}) since they aim to keep the image background unaltered.  
\subsubsection{InstaGAN}
\label{sec:InstaGAN}
Solutions exist to 
to deal with datasets that exhibit more evident changes in the form of objects or include multiple instances of objects. To this end, InstaGAN \citep{Sangwoo2018} requires information on the instances to be modified, obtained through segmentation masks a priori provided: $S$ and $T$ must contain both images and masks for all elements of interest. 
 A set of generators supplements the original generator: 
each generator operates on a segmentation mask and produces a translated mask in the target domain.
The cross-entropy used in CGAN 
is modified by adding weighted terms that include the cycle-consistency loss and a background-preserving loss. 
InstaGAN produces high-quality images where only selected regions are modified. 
However, it adds a new constraint: it requires a segmentation mask associated with each object. Once again, this is problematic for image culturization based on custom-built datasets composed of images downloaded from the web. 

\subsubsection{UNIT}
\label{sec:UNIT}
Approaches exist that emphasize the dichotomy between high-level semantics (e.g., depending on the object's functionalities or affordances) and low-level features (e.g., ``brushstrokes of cultural features").
UNsupervised Image-to-image Translation (UNIT) \citep{Unit2017} is a representative of this class based on CoGAN \citep{Ming2016}. CoGAN consists of two GANs: each GAN takes a random vector as input without being conditioned to any input image. Generators are trained to produce pairs of images in the two domains $S$ and $T$; discriminators are trained to distinguish generated images from real ones. Very importantly, the weights of the first few layers of the generators and those of the last few layers of the discriminators (responsible for decoding and encoding high-level semantics) are shared. Thanks to this, CoGAN generates pairs of related images in $S$ and $T$ with the same high-level features but different low-level ones. 
However, CoGAN is unsuitable for image-to-image translation because it is fed with random input $z$. UNIT makes the network conditioned by adding variational autoencoders  \citep{VAeGAN}, trained to encode images from the two domains  $S$ and $T$ into a shared latent code $z$, which is then fed to a CoGAN-like network to produce images in  $S$ and $T$. This process 
implies a cycle-consistency constraint comparable with CycleGAN's.

\section{Materials and Methods}
\label{Materials and Methods}

This section describes the process to culturize images 
and 
to explore our research question 
with recruited participants.

\subsection{Choosing a solution for GAN-based culturization}
\label{Solution}

The first step is the choice of an appropriate model for image culturization. To this end,  we considered all the solutions mentioned in Sections from 
\ref{sec:Pix2Pix} to \ref{sec:UNIT}: Pix2Pix, CycleGAN, Attention-guided GAN, InstaGAN, UNIT.

GANs evaluation \citep{Proecons2018} is a challenging task. GANs are trained to reach an equilibrium situation where the generator can ``cheat" the discriminator. Still, the former, taken alone, is not associated with a cost function to be minimized. Under these conditions, it is hard to predict when the generator will produce samples that fit the target probability distribution at their best. Then, researchers have defined qualitative and quantitative tools to evaluate GANs. 
Qualitative assessment of the generated images is often performed 
using crowdsourcing platforms such as Amazon Mechanical Turk \citep{Isola2017, Jun-Yan2017, Zili2017, Mejjati2018}.
Quantitative evaluation can be performed using such metrics as the Inception score (IS) \citep{Tim2016} and the Fréchet Inception Distance (FID) \citep{FID2017} -- or other metrics \citep{pang2021imagetoimage}.
IS evaluates generated images based on how Inception v3, a widely-used image recognition model that attains greater than 78.1\% accuracy on the ImageNet \citep{Szegedy20162818}, classifies them. 
In doing this, IS measures if there is sufficient diversity among the generated samples, addressing the so-called problem of the ``collapse of the model": the generator learns to generate a specific image of the target domain and keeps on generating the same image (even when varying the input) because the latter is very good in deceiving the discriminator.
FID also uses a pre-trained Inception v3 model to measure the quality of the generated images. Still, it does so differently, i.e., by evaluating the accuracy of the Inception v3 model in classifying GAN-generated images compared with real ones. Consequently, FID better captures the similarity of generated images to real ones than IS. 

In principle, we might compare the solutions for image culturization discussed in the previous section using FID: this is done, for example, in \citep{Zhao2020}, which unfortunately shows how FID values can significantly vary depending on specific image-to-image translation tasks. Therefore, following the tradition of qualitative evaluation \citep{Isola2017, Jun-Yan2017, Zili2017, Mejjati2018}, we found it more appropriate to subjectively compare the results of different GANs  (FID will play a role in the next section in tuning hyperparameters of the selected GAN solution). 

Specifically, we observed that Pix2Pix and similar approaches require paired datasets, which is a limitation since image culturization should work on custom-built datasets, e.g., composed of images downloaded from the web. InstaGAN did not reveal a feasible solution for similar reasons since it requires a carefully designed dataset composed of images belonging to different cultural domains and the corresponding masks, even if it might be the optimal choice to alter the form of multiple objects in the same image. 
Following these initial considerations, we implemented CycleGAN, AttentionGAN, and UNIT, which do not pose constraints on the dataset. Then, volunteers visually inspected the results to assess how each solution can meet the requirements of altering objects' visual appearance (color, material, pattern, and form) for image culturization. Tests confirmed that AttentionGAN has the strongest limitations when an image-to-image translation requires altering the object form, which is often needed in image culturization (see the vases in the first row of Figure \ref{all}). 
Both CycleGAN and UNIT might be good choices, as they are sufficiently capable of altering all visual features. After qualitative evaluation, and since CycleGAN is the basis for many other approaches (including InstaGAN that might be reconsidered in the future), we finally selected it as the best candidate for image culturization. However, we do not see any obstacles in implementing the same procedure described in the next Sections with a different GAN. 



\subsection{Culturization pipeline}
\label{sec:cult-pipeline}

In the following, we refer to $S$ as the set of images defining a start domain (e.g., Greek vases), $T$ as the set of images defining a target domain (e.g., Chinese vases), $S\to T$ as the images generated starting from $S$ (e.g., Greek vases changed to Chinese style), $T \to S$ as the images generated starting from $T$ (e.g., Chinese vases changed to Greek style). 

\begin{figure}
    \centering
    \includegraphics[width=\columnwidth]{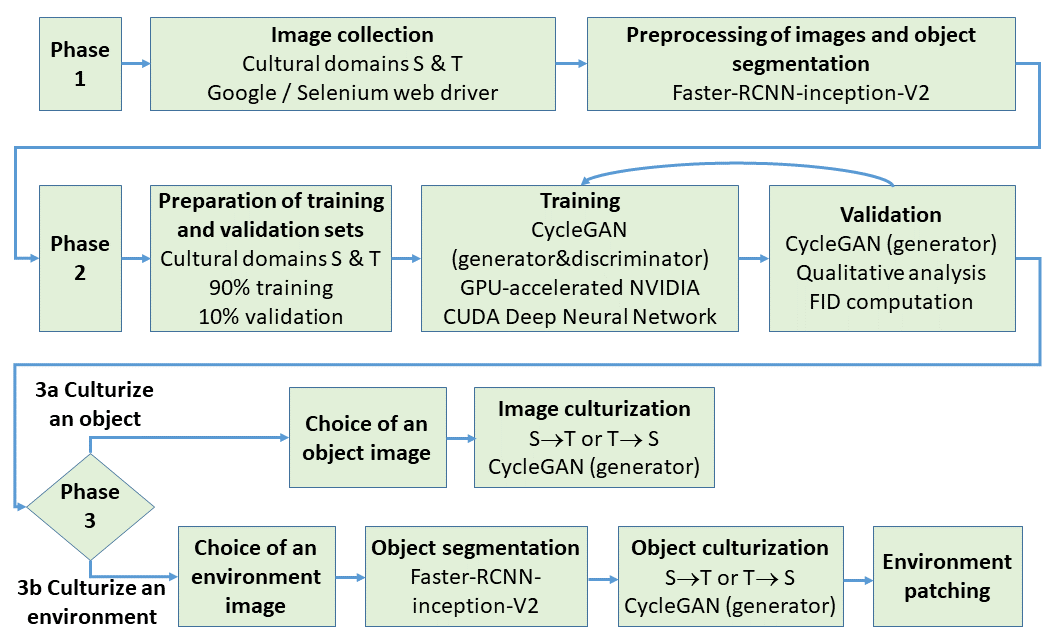}
    \caption{Culturization process.}
    \label{fig:cultu}
\end{figure}

The culturization process includes three phases, Figure \ref{fig:cultu}: the building of datasets (phase 1), training and validation (phase 2), and the production of culturized images for interaction with people (phase 3). We did not follow the usual training-validation-test pipeline since we are only interested in producing images of sufficient quality for evaluation with people, not in assessing how the CycleGAN model performs on a holdout test set. Additionally, following a preliminary investigation, we realized that the default hyperparameters in the CycleGAN implementation\footnote{Available at github.com/junyanz/pytorch-CycleGAN-and-pix2pix} are appropriate to produce images of sufficient quality (which is good because retraining the model takes tens of hours with our hardware/software configuration). Therefore, after collecting datasets describing a given object class (e.g., vases) in the source and target cultural domains $S$ and $T$ (e.g., Greek and Japanese), we split them into a training and validation set. Then, we performed training and validation a few times to tune the number of training epochs and the dimensions of the datasets based on a qualitative and quantitative evaluation of the culturized images. 



\subsubsection{Phase 1}
\label{sec:Phase 1}
In phase 1, two fully automated steps are required by the system to build datasets, managed by a script in Python.
\begin{itemize}
    \item \textit{Image collection}. The script requires to enter the name of an object class (e.g., vase, sofa, pillow, lamp, etc.), a cultural domain (e.g., Greek, Indian, Moroccan, Chinese, etc.), 
and the number of images to download. The images are automatically downloaded from Google through the Selenium Web Driver API, a collection of open-source APIs to automate the testing of a web application. 

\item \textit{Preprocessing of images and object segmentation}. Downloaded images 
are preprocessed to make them suitable for CycleGAN: first, images are resized to the required $256\times 256$ dimensions; then, objects are segmented using a network capable of identifying the desired objects within the image. 
Object detection and segmentation rely on 
the \textit{Faster-RCNN-inception-V2} pre-trained network. To this end, the candidate objects to extract need to be chosen, 
and networks are trained by labeling around $3,000$ images via the labelImg tool. 
If multiple instances of an object are present, they are all detected and segmented.
\end{itemize}

\begin{remark}
\label{R2}
The first images returned by Google are usually more appropriate to represent the domain than the last ones. 
Therefore, the script downloads only a small number of images returned by the initial Google search and then searches for correlated pages until it produces the required sample. Downloaded images are manually checked and possibly discarded if needed to obtain better results.
\end{remark}

\subsubsection{Phase 2}
\label{sec:Phase 2}
In phase 2, three steps are required to train and validate the model to map images from $S$ to $T$ and vice versa. 

\begin{itemize}
\item \textit{Preparation of training and validation sets}. We consider two datasets of images $S$ and $T$ 
corresponding to the same object class in the two cultural domains 
and split them into a training set (90\% of the dataset, built with an equal number of images from $S$ and $T$) and a validation set (10\% of the dataset). 
\item \textit{Training}. The network is trained to generate images $S \to T$ starting from $S$ as well as images $T \to S$ starting from $T$. 
We train the network with a GPU TESLA P100 using the GPU-accelerated NVIDIA CUDA Deep Neural Network library (about 30 hours for 200 epochs).     

\item \textit{Validation}. The validation set is used to generate images $S \to T$ starting from $S$ as well as images $T \to S$ starting from $T$. Only the trained generator is used, whereas the discriminator is no longer needed. The resulting images $S \to T$ and $T \to S$ undergo a qualitative analysis to visually judge their quality and the FID score is computed 
to confirm that the results are acceptable. 
If needed, the network is trained again by changing the number of training epochs and the dimensions of the dataset. 
\end{itemize}

During training and validation, 
we did not find a significant difference in the perceived quality and FID values when changing the number of training epochs from 150 to 400 (training time increasing from 20h to 50h and more). For example, when processing images of sofas, FID values ranged from 129.18 to 122.33  (from Indian to classic European) by increasing the number of epochs; when processing images of pillows, FID values ranged from 78.60 to 83.02 (from classic European to Indian).
It must also be remembered that the range of FID values depends on the application for which GANs are used: FID is typically used to compare different approaches rather than providing an absolute estimate of the quality of the process. The work in \citep{Zhao2020}, which focuses on human faces, reports FID values ranging from 23.72 to 48.71 for ``glass removal", from 16.63 to 36.17 for ``male to female", and from 93.58 to 102.92 for ``selfie to anime". Since CycleGAN has limitations in modifying the form of objects (even if it is more performing than AttentionGAN), FID values tend to be quite high in our case. However, they do not change significantly with the number of training epochs. This result is also confirmed by the qualitative evaluation of culturized images, which motivated us to keep the default value of 200 training epochs in the chosen CycleGAN implementation. The same rationale motivated us to use training datasets of 1,000 images, which allows for a good compromise between FID values, perceived quality, and variability.




\subsubsection{Phase 3}
\label{sec:Phase 3}
In phase 3, the paths in Figure \ref{fig:cultu} differ depending on whether one wants to culturize individual objects by ignoring the background or multiple objects that in a complex environment.

In the first case (3a), it is sufficient to feed CycleGAN with an image belonging to $S$ or $T$.
\begin{itemize}
    \item \textit{Choice of an object image}. The image of an object belonging to a cultural domain $S$ or $T$ is provided.
    \item \textit{Image culturization}. The image is translated from $S$ to $T$ or $T$ to $S$ using the CycleGAN generator. 
\end{itemize}
 
In the second case (3b), if the objects to be culturized are part of a complex environment, and the environment itself matters (as in Figure \ref{fig:cleaning}), four steps are required. 

\begin{itemize}
    \item \textit{Choice of an environment image}. The image of an environment belonging to a domain $S$ or $T$ is provided.
    \item \textit{Object segmentation}. The image of the environment is preprocessed using \textit{Faster-RCNN-inception-V2} to extract all objects that are a candidate for culturization. 
    \item \textit{Object culturization}. CycleGAN generator is used to map each extracted object from the domain $S$ to $T$ or $T$ to $S$.  
    \item \textit{Environment patching}. The original image is patched with the culturized objects. Currently, this step is not automated but manually performed with a tool. 
\end{itemize}
    

The images produced can finally be used for culturally-competent real-time interaction: as an example, think about the personal assistant Tetsuwan Atomu presented in the short Yukiko story that opened the article. 
However, culturally competent real-time interaction is not the focus of the experiments performed in this work. Due to the novelty of the culturization concept, we decided to explore our research question through an online questionnaire to measure the impact of image culturization on people, which allows for the involvement of a greater number of recruited participants.

%
\subsection{Study design}
\label{Experiments}

\subsubsection{Research question and general considerations}
\label{General considerations}
As anticipated in the introduction, we aimed to explore the following research question:

\begin{quote}
`to what extent do people appreciate images that were culturized to make them coherent with their culture?'   
\end{quote}

The general idea was to show participants images of different objects and environments belonging to European and non-European cultural domains and ask them some questions. To this end, we decided to do experiments with Italian participants only to simplify recruitment and because it is easier for us to guess whether objects will look familiar to Italians or not (since all authors are Italian, trying to select objects that look familiar to Indians or Moroccans incurs a higher risk of stereotyping). 

To this end, we considered four object classes (vases, sofas, pillows, and lamps) and, for each class, four sets of images produced in different ways: one set was downloaded from the Internet as belonging to the European culture and tradition ($E$); one was downloaded from the Internet as belonging to a different culture and tradition (Chinese, Indian or Moroccan, $O$ for other); one was a non-European object culturized using GANs ($O \to E$); one was a culturized European object  ($E \to O$). 

\begin{remark} 
\label{R5}
A limitation of this approach, as discussed in section \ref{Introduction}, is that we focussed on the culturization of individual objects instead of complete scenes. However, objects play a crucial role in conveying cultural meanings and can strongly influence how an image is perceived within a specific cultural context. Understanding and addressing object-level translation is a vital step toward achieving more comprehensive scene-level translation in future research. In the questionnaire, we selected four object classes (vases, sofas, pillows, and lamps) because they exhibit sufficient diversity and are culturally characterized in terms of color, material, pattern, and form, see Remark \ref{R1}. Since we did not perform any formal analysis to determine what objects are better for making cultural differences emerge, we cannot exclude that the study would produce different results when culturizing different objects. 
\end{remark}

\begin{remark} 
\label{R3}
Considering Italian participants only is another limitation, as discussed in \ref{Introduction}. However, notice that, unlike other studies, the independent variable will be the cultural context of the object (not the participants), and the dependent variable will be the score assigned by participants. Under this aspect, the study seems correct, even if the results can only be generalized to Italian people. Moreover, we think that there are unique aspects of Italian culture that make it an interesting and relevant case study for exploring the impact of image culturization. Italy has shown to be a culturally conservative society, with a lesser openness to novelty \citep{Schwartz19921, Rubera2011459}, and Italians tend to have specific sensitivities and expectations regarding the accurate representation of their cultural identity and traditions, which can shape individuals' preferences and inclinations toward certain visual styles.  
Since the principles underpinning this research are general, future research can focus on different cultural contexts to validate and extend findings across diverse cultural groups. 
\end{remark}

\begin{remark}
\label{R4}
Trying to define a ``European" cultural style $E$ may seem like an ill-posed problem since Europe is a collection of diverse nations and cultural groups. The same can be said for the ``other" style $O$, which we define as ``not $E$." However, the rest of this work does not rely on the existence of an a priori definition of $E$ but rather pursues a pragmatic approach. Since we conjecture that Italians (and Europeans) have a subjective perception of what is $E$ and what is $O$, the initial questions will be devoted to evaluating Hypothesis 1 below, i.e., whether objects are recognized by participants as belonging to $E$ or $O$. 
\end{remark}

\subsubsection{Hypotheses}
\label{sec:Hypotheses}
To explore our research question, we searched for evidence supporting the four hypotheses below. 

\begin{hypothesis}Objects are correctly recognized by Italian participants as belonging to the $E$ culture. This hypothesis holds both when the object originally belonged to the $E$ culture and when it was culturized from  $O$ to $E$.
\end{hypothesis}
 \begin{hypothesis}European objects are generally preferred by Italian participants. This hypothesis holds both when the object originally belonged to the $E$ culture and when it was culturized from  $O$ to $E$.
 \end{hypothesis}
\begin{hypothesis} Italian participants who prefer $E$ objects also tend to prefer objects culturized from $O$ to $E$.
 \end{hypothesis}
 \begin{hypothesis} Italian participants perceive environments as more realistic when objects in the image were segmented, culturized with GANs, and then re-inserted in the image rather than when we download new objects from the Internet, clip them, and overlay them on the background.
 \end{hypothesis}

Hypothesis 1, as anticipated, aims to confirm that participants have a shared perception of what is $E$ and what is $O$, laying the basis for the following hypotheses. Concerning Hypothesis 2, each person has their preferences: hypothesizing that Italian people will prefer European objects incurs the risk of stereotyping and, generally speaking, is not true. Different persons may be more or less attracted by objects belonging to their own or other cultures, which also depends on individual objects. For example, the same person may adore colorful Moroccan lamps and, at the same time, be particularly attracted by the design of a lamp produced in Sweden or France. We are aware of this and do not intend to make stereotyped claims. The objective of Hypothesis 2, together with Hypothesis 3, is to explore if there are some people (not necessarily all people) with whom using images that match their cultural background can be a winning strategy. Hypothesis 4 aims to confirm the approach's feasibility when considering complex environments where individual objects need to be culturized.  


\begin{figure}
  \centering
  \includegraphics[width=\linewidth]{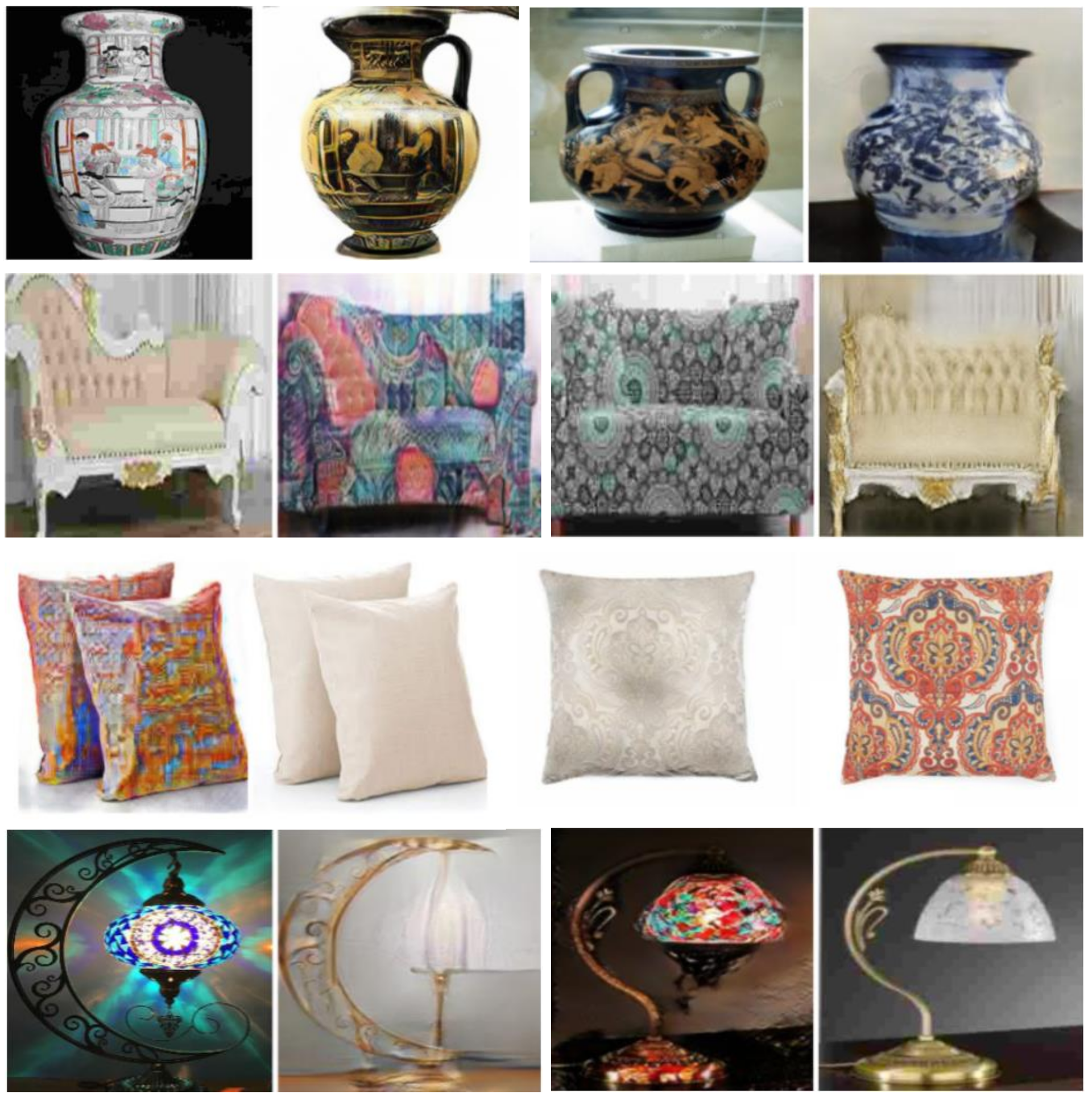}
  \caption{Pairwise object comparisons. $E$: downloaded from the Internet as belonging to the European culture; $O$: downloaded as belonging to a different culture; $O \to E$: non-European object culturized using GANs; $E \to O$: European object culturized using GANs. 1st row: $O$ vs. $O \to E$;  $E$ vs. $E \to O$. 2nd row: $E$ vs. $E \to O$; $O$ vs. $O \to E$; 3rd row: $E \to O$ vs. $E$; $O \to E$ vs. $O$. 4th row: $O$ vs. $O \to E$; $E \to O$ vs. $E$.}
  \label{all}
\end{figure}

\subsubsection{Questionnaire items}
\label{sec:Questionnaire items}
After preparing images using the procedures in section \ref{sec:cult-pipeline}, we set up a Google Form questionnaire. The questionnaire was anonymous and included the four sections in Figure \ref{Tableq}, showing pictures and asking related questions\footnote{Full questionnaire available at 
 {https://bit.ly/3zjVHWL}
 }.



\begin{table}
\begin{tabular} {ll}
\hline
First Section: Personal questions & \\
\hline
Second Section: Recognizing culture and expressing preferences (5-points Likert scale) & \\
Q1: How close is this object to European culture and tradition? & (H1) \\
Q2: Do you like this object? & (H2, H3) \\
 \hspace{1cm} 4 Vases (1 $E$; 1  $O\to E$; 1 $E\to O$; 1 $O$) & \\
 \hspace{1cm} 4 Sofas (1 $E$; 1  $O\to E$; 1 $E\to O$; 1 $O$) & \\
 \hspace{1cm} 4 Pillows (1 $E$; 1  $O\to E$; 1 $E\to O$; 1 $O$) & \\
 \hspace{1cm} 4 Lamps (1 $E$; 1  $O\to E$; 1 $E\to O$; 1 $O$) & \\
\hline
Third Section: Comparing culturized and non-culturized objects (binary choice) & \\
Q1: Which object best represents European culture and tradition? & (H1) \\
Q2: Which object do you like the most? & (H2, H3) \\
 \hspace{1cm} 4 Vase comparisons (2 $E$ vs. $E\to O$; 2 $O$ vs. $O\to E$) & \\
 \hspace{1cm} 4 Sofa comparisons (2 $E$ vs. $E\to O$; 2 $O$ vs. $O\to E$) & \\
 \hspace{1cm} 4 Pillow comparisons (2 $E$ vs. $E\to O$; 2 $O$ vs. $O\to E$) & \\
 \hspace{1cm} 4 Lamp comparisons (2 $E$ vs. $E\to O$; 2 $O$ vs. $O\to E$) & \\
 \hline
Fourth Section: Comparing the realism of environments (binary choice) & \\
Q1: Which image look more realistic? & (H4) \\
\hspace{1cm} 8 Environment comparisons (8 $Gcult$ vs. $Pcult$) & \\
\hline 
\end{tabular}
\caption{Questionnaire's structure: after \textit{Personal questions}, 16 objects are shown in the section \textit{Recognizing culture and expressing preferences}, questions Q1 and Q2; 16 pairwise object comparisons are shown in the section \textit{Comparing culturized and non-culturized objects}, questions Q1 and Q2; 8 pairwise environment comparisons are shown in the section \textit{Comparing the realism of environments}, question Q1. We use question Q1 in the second and third sections to evaluate Hypothesis 1, Q2 in the second and third sections to evaluate Hypotheses 2 and 3, and Q1 in the fourth section to evaluate Hypotheses 4.}

\label{Tableq}
\end{table}

\begin{figure}
  \centering
  \includegraphics[width=0.8\linewidth]{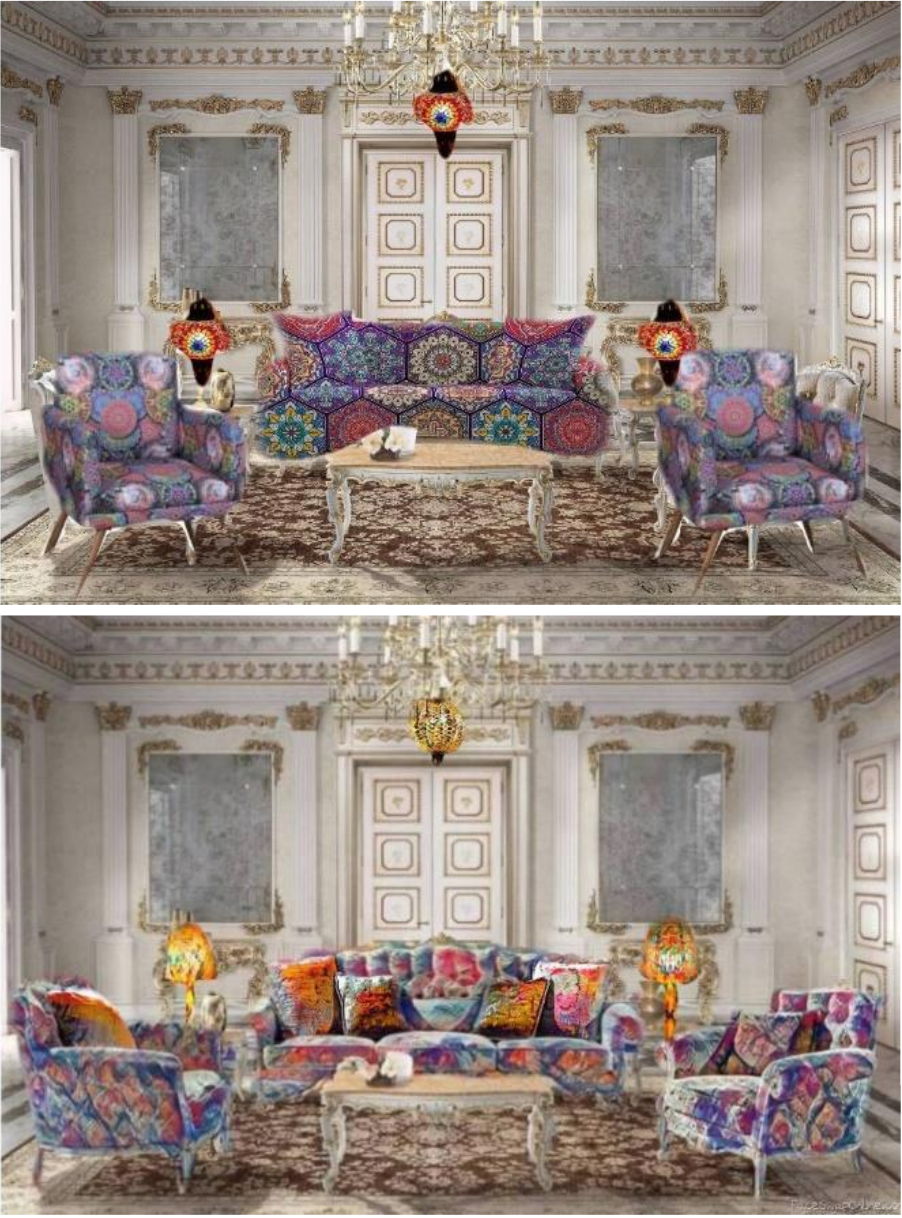}
  \caption{Top: $Pcult$ image patched with objects downloaded from the Internet, clipped and overlayed on the original image; Bottom: $Gcult$ image patched with GAN-modified objects. 
  }
  \label{rooms}
\end{figure}

\begin{enumerate}
\item \textit{Personal questions}: the respondents had to declare their age (\textit{less than 20}, \textit{20--29}, \textit{30--39}, \textit{40--49}, \textit{50--59}, \textit{60--69}, \textit{70 and over}) and if how often travel out of Italy (\textit{never or very rarely}; \textit{yes, but mostly in Europe}; \textit{yes, both in Europe and outside Europe});
\item \textit{Recognizing culture and expressing preferences}: we showed respondents four vases, four sofas, four pillows, and four lamps. For each image, the respondents had to reply to the following two questions by assigning a score on a 5-point Likert scale. Q1: `How close is this object to European culture and tradition?' (from \textit{1:very far} to \textit{5:very close}); Q2: `Do you like this object?' (from \textit{1:not at all} to \textit{5: a lot}). For each object class (vases, sofas, pillows, and lamps), the four objects shown to the respondent belonged to different groups ($E$, $O \to E$, $E \to O$, and $O$). 
\item \textit{Comparing culturized and non-culturized objects}: we showed respondents 16 pairwise comparisons, four of which concerned vases, four sofas, four pillows, and four lamps. For each comparison, two objects were shown side by side: the original one (which can either be $E$ or $O$) and its corresponding GAN counterpart ($E \to O$ or $O \to E$, depending on the original one). The respondents had to reply to the following two questions. Q1: `Which object best represents European culture and tradition?' (\textit{the one on the left} or \textit{the one on the right}); Q2: `Which object do you like the most?' (\textit{the one on the left} or \textit{the one on the right}). We assigned a score of 1 to the winning object and 0 to the loser. The order in which original and modified objects appeared in the pair varied for each comparison. Figure \ref{all} shows some of the pairs presented to respondents. 
\item \textit{Comparing the realism of environments}: we showed respondents 8 pairwise comparisons. For each comparison, two environments were presented to the respondent: an environment that was modified by segmenting objects, 
culturizing them with GANs, and then re-inserting them into the original image ($Gcult$); 
an environment modified by substituting the original objects with objects of different cultures downloaded from the Internet, clipped, and overlayed on the original image ($PCult$). The respondent had to reply to the following question. Q1: `Which image looks more realistic?' (\textit{the one on the top} or \textit{the one on the bottom}), Figure \ref{rooms}. We assigned a score of 1 to the winning environment and 0 to the loser.
\end{enumerate}

We use question Q1 in the questionnaire's sections 2 and 3 to evaluate Hypothesis 1, Q2 in sections 2 and 3 to evaluate Hypotheses 2 and 3, and Q1 in section 4 to evaluate Hypotheses 4.

Note that we chose objects whose cultural belonging was emphasized by the color, material, pattern, and form, Remark \ref{R5}. Greek vases were easily recognizable as belonging to the European culture; non-European vases were mostly Chinese and Japanese. Figure \ref{all} shows that, when culturizing vases through GANs, the original Chinese or Greek drawings are recognizable in their GAN-modified counterparts: a Chinese vase with greek warriors and a Greek vase with an oriental tavern can be spotted in the Figure. 
Similarly, we chose European sofas designed in classic style. One could object that this kind of sofa is not common in ordinary European houses: however, we wanted to avoid international, modern-style sofas that may not be immediately recognizable as European. Pillows were probably less recognizable as 
European / non-European 
since colorful pillows are customary also in ordinary European houses. Finally, the cultural belonging of lamps is evident: see the moon-shaped Moroccan lamp versus its European counterpart and the European ``night table" lamp rethought in a Moroccan style.

\subsubsection{Data collection}
\label{sec:Data collection}
The Google Form was online from 18/09/2021 to 7/10/2021. We recruited participants through the University of Genova's public social networks (Facebook and Twitter) and student networks. The questionnaire included the following introduction in Italian: `The questionnaire will show you a sequence of images and a few questions. The images show different objects and environments associated with different cultures. We will ask you to rate how close the images are to your culture and express your tastes and preferences. Please look at the images and answer truthfully based on your feelings: some questions require you to respond with a score between 1 and 5, and others require you to choose between two images. The questionnaire should take between 10 and 20 minutes.'

\section{Results}
\label{results}

This section describes collected data and related analyses. Note that both the second and third sections of the questionnaire in Table \ref{Tableq} evaluate Hypotheses 1, 2, and 3, even if through different strategies, allowing us to compare results and possibly strengthen conclusions. The fourth section evaluates Hypothesis 4. 

\subsection{First Section: Personal questions}
\label{Anagraphic data}

Overall, N=392 participants filled the questionnaire, out of which: 7.9\% were less than 20 years old; 21.9\% were in the range 20-29; 13.8\% in the range 30-39; 20.4\% in the range 40-49; 20.9\% in the range 50-59; 11.5\% in the range 60-60; 3.6\% are 70 years or older. Concerning travels, 26\% of the participants declared they never travel out of Italy or very rarely; 48.5\% frequently travel out of Italy, but mostly in Europe; 25.5\% travel both in Europe and out of Europe.

\subsection{Second Section: Recognizing culture and expressing preferences}
\label{Recognizing culture and expressing preferences}
Table \ref{Table1} shows the results of the second section of the questionnaire. For each question Q1 and Q2, the table reports the average score and its standard deviation obtained by objects (vase, sofa, pillow, or lamp, both individually taken and all together) belonging to different groups ($E$, $O \to E$, $E \to O$, $O$), computed over N respondents.  

\begin{table}
\begin{tabular}{ |c c|c c|c c|c c|c c| }
\hline
N=392 &  & Q1 & Q2 & Q1 & Q2 & Q1 & Q2 & Q1 & Q2 \\ \hline
 \multicolumn{2}{|c|}{ } &
 \multicolumn{2}{|c|}{$E$ }&
 \multicolumn{2}{|c|}{$O\to E$ }&
 \multicolumn{2}{|c|}{$E\to O$ } &
 \multicolumn{2}{|c|}{$O$ }\\
 \hline
Vases & av. & 4.54 & 4.04 & 4.00 & 3.66 & 2.49 & 2.23 & 2.53 & 2.52 \\
 & std. & 0.95 & 1.02 & 1.28 & 1.12 & 1.01 & 1.10 & 1.08 & 1.22 \\
 \hline
 Sofas & av. & 4.62 & 4.06 & 3.67 & 2.16 & 2.68 & 2.01 & 2.62 & 2.56 \\
 & std. & 0.69 & 0.87 & 1.15 & 1.10 & 1.11 & 1.19 & 1.12 & 1.34 \\
  \hline
   Pillows & av. & 4.23 & 3.27 & 2.85 & 2.31 & 2.49 & 2.66 & 2.50 & 2.75 \\
 & std. & 0.91 & 1.27 & 1.01 & 1.11 & 1.03 & 1.29 & 1.07 & 1.25 \\
 \hline
    Lamps & av. & 4.41 & 2.79 & 2.68 & 2.23 & 1.39 & 2.34 & 2.42 & 2.80 \\
 & std. & 0.75 & 1.14 & 1.05 & 1.19 & 0.77 & 1.22 & 1.07 & 1.30 \\ \hline
    All & av. & 4.45 & 3.54 & 3.30 & 2.59 & 2.26 & 2.31 & 2.52 & 2.66 \\
 & std. & 0.52 & 0.67 & 0.64 & 0.70 & 0.63 & 0.84 & 0.66 & 0.87 \\
  \hline
\end{tabular}
  \caption{Questionnaire's second section. For each question Q1 and Q2, the table reports the average score and its standard deviation obtained by objects (vase, sofa, pillow, or lamp, both individually taken and all together) belonging to different groups ($E$, $O \to E$, $E \to O$, $O$), N=392 respondents.}
  \label{Table1}
\end{table}



\begin{table}
\begin{tabular}{ |c c|c c|c c|c c|c c| }
\hline
N=117 &  & Q1 & Q2 & Q1 & Q2 & Q1 & Q2 & Q1 & Q2 \\ \hline
 \multicolumn{2}{|c|}{ } &
 \multicolumn{2}{|c|}{$E$ }&
 \multicolumn{2}{|c|}{$O\to E$ }&
 \multicolumn{2}{|c|}{$E\to O$ } &
 \multicolumn{2}{|c|}{$O$ }\\
 \hline
Vases & av. & 4.48 & 4.03 & 4.03 & 3.59 & 2.38 & 2.10 & 2.83 & 2.66 \\
 & std. & 0.92 & 0.98 & 1.26 & 1.17 & 0.84 & 1.07 & 0.92 & 1.23 \\
 \hline
 Sofas & av. & 4.51 & 4.21 & 3.93 & 2.44 & 2.76 & 2.00 & 2.50 & 2.18 \\
 & std. & 0.78 & 0.79 & 1.05 & 1.11 & 1.15 & 1.23 & 1.06 & 1.14 \\
  \hline
   Pillows & av. & 4.05 & 3.27 & 2.82 & 2.00 & 2.27 & 2.56 & 2.31 & 2.61 \\
 & std. & 1.01 & 1.22 & 1.04 & 1.02 & 0.97 & 1.26 & 0.96 & 1.20 \\
 \hline
    Lamps & av. & 4.35 & 2.97 & 2.76 & 2.50 & 1.21 & 2.79 & 2.20 & 2.87 \\
 & std. & 0.77 & 1.12 & 1.05 & 1.17 & 0.55 & 1.33 & 0.92 & 1.34 \\ \hline
    All & av. & 4.35 & 3.62 & 3.38 & 2.63 & 2.16 & 2.37 & 2.46 & 2.58 \\
 & std. & 0.57 & 0.62 & 0.65 & 0.66 & 0.55 & 0.85 & 0.61 & 0.82 \\
  \hline
\end{tabular}
  \caption{Questionnaire's second section, age$< 30$, N=117. Data are organized as in Table \ref{Table1}}
  \label{Table2}
\end{table}

\subsubsection{Evaluating Hypothesis 1}
\label{Evaluating Hypothesis 1}

To search for evidence supporting Hypothesis 1, we considered Q1: `How close is this object to European culture and tradition?' Specifically, we 
explored the relation between the mean score $\mu_E$ or $\mu_{O \to E}$ assigned to $E$ or $O \to E$ objects with the mean score of other classes.   
To this end 
we performed an ANOVA test with $\alpha=0.05$ by analyzing all groups together ($E$, $O \to E$, $E \to O$, $O$) and a Tukey HSD test to check if pairwise groups are statistically different. 
Due to the sample size N=392, we did not need to check distribution normality.

Table \ref{Table1} shows that concerning Q1, 
\begin{itemize}
\item for any object class, $E$ objects achieved the highest average score;
\item for any object class, $O \to E$ objects achieved the second-highest average score.
\end{itemize}

The ANOVA returned that we can reject the null hypothesis (i.e., the hypothesis that differences between classes are due to chance). When pairwise taken, the Tukey HSD test returned that all groups' averages are significantly different with $p<0.05$ except for the couple ($E \to O$, $O$) in the vases, sofas, and pillows classes. 


In summarizing, with a focus on $\mu_{E}$ and $\mu_{O \to E}$ in Q1,
\begin{itemize}
\item $\mu_{E}>\mu_{O \to E} > \mu_{E \to O}=\mu_{O}$ for vases, sofas, pillows,
\item $\mu_{E}>\mu_{O \to E} > \mu_{O}>\mu_{E \to O}$ for lamps,
\item $\mu_{E}>\mu_{O \to E} > \mu_{O}>\mu_{E \to O}$  when considering all classes together,
\end{itemize} 
which supports Hypothesis 1 since $E$ and $O\to E$ classes are recognized as more ``European" than others.

\subsubsection{Evaluating Hypothesis 2}
\label{Evaluating Hypothesis 2}

To search for evidence supporting Hypothesis 2, we considered Q2: `Do you like this object?' by following the same approach described in the previous section for statistical analysis. 

Table \ref{Table1} shows that concerning Q2, 
\begin{itemize}
\item for any object class, $E$ objects achieved the highest average score;
\item for vases, the second-highest average score was achieved by $O \to E$ objects; 
\end{itemize}

The ANOVA returned that we can reject the null hypothesis. The Tukey HSD test returned that all groups' averages, when pairwise taken, are significantly different with $p<0.05$, with the only exception of the couple ($O \to E$, $E \to O$) in the sofa class, ($E \to O$, $O$) in the pillow class, ($E$, $O$), ($O \to E$, $E \to O$) in the lamp class, and ($O \to E$, $O$) in all classes taken together. 


In summarizing, with a focus on $\mu_{E}$ and $\mu_{O \to E}$ in Q2,
\begin{itemize}
\item $\mu_{E}>\mu_{O \to E} > \mu_{O}>\mu_{E \to O}$ for vases,
\item $\mu_{E}>\mu_{O} > \mu_{O \to E}=\mu_{E \to O}$ for sofas,
\item $\mu_{E}>\mu_{E \to O} = \mu_{O} > \mu_{O \to E}$ for pillows,
\item $\mu_{E}=\mu_{O} > \mu_{O \to E} = \mu_{E \to O}$ for lamps, 
\item $\mu_{E}> \mu_{O \to E}=\mu_{O} >\mu_{E \to O}$ when considering all classes together, 
\end{itemize}
which supports Hypothesis 2 only partially because, in several cases, $E\to O$ and $O$ objects (which should appear less ``European" according to Hypothesis 1) are preferred to $O \to E$ ones (more ``European"). 
Generally speaking, and according to expectations, Italian participants exhibit a conservative preference for familiar $E$ objects \citep{Rubera2011459}.

\subsubsection{Stratified analysis}
\label{Evaluating Hypotheses 2 and 3: stratified analysis}
We performed a stratified analysis controlling for age and travel habits. Despite minor differences, the results were similar. However, something interesting emerged when considering only respondents with age $< 30$ (N=117), Table \ref{Table2}.
 
Concerning Hypothesis 1, results were mostly confirmed: the statistical analyses (ANOVA and Tukey HSD,  $p<0.05$) showed that, for any object individually taken and for all objects taken together, it always holds $\mu_{E}>\mu_{O \to E} > \mu_{O}$ with $p<0.05$. 
However, concerning Hypothesis 2, there are some differences with the case in which we considered all age ranges. By focussing on $E$ and $O\to E$, the statistical analysis on vases returned, as previously, $\mu_{E}>\mu_{O \to E} > \mu_{O}$ with $p<0.05$; the analysis on sofas 
returned, in this case, $\mu_{E}>\mu_{O \to E} > \mu_{O}$ with $p<0.05$, supporting Hypothesis 2; the analysis on all objects taken together returned $\mu_{E}>\mu_{O \to E} = \mu_{O}$ as in the previous case, but now the average score of $O \to E$ is slightly higher than 
$O$, even if the difference is not statistically significant with $p<0.05$. Results support Hypothesis 2 to a higher degree, but still partially.  

\subsubsection{Evaluating Hypothesis 3}
\label{Evaluating Hypothesis 3}
To search for evidence supporting Hypothesis 3, we considered Q2: `Do you like this object?' for the whole sample with N=391. Specifically, we considered the N participants and the four groups ($E$, $O \to E$, $E \to O$, and $O$). Next,  for each participant and group, we computed the average score for all objects belonging to that group: this process yielded four N-sized vectors $A_{E}$, $A_{O\to E}$, $A_{E\to O}$, $A_{O}$, one per group. The N elements of a vector represent the average scores assigned to that group by the N participants.  

Finally, we computed Pearson's $\rho$ to correlate $A_{E}$ with $A_{O\to E}$, $A_{E\to O}$, and $A_{O}$: $\rho$ ranges from $-1$ to $1$ and a higher correlation $\rho_{E, O \to E}$ (respectively, $\rho_{E, E \to O}$, $\rho_{E, O}$) means that respondents giving high scores to $E$ objects tend to give high scores to $O \to E$ objects as well (respectively, $E \to O$, and $O$). The analysis returned $\rho_{E, O \to E}=0.48$, $\rho_{E, E \to O}=0.16$, $\rho_{E, O}=0.15$ supporting Hypothesis 3: respondents with a preference for $E$ objects tend to prefer $O \to E$ objects to other groups (i.e., participants are coherent in preferring objects that seem or does not seem ``European", even when the European look is achieved through culturization). 

\begin{table}
\begin{tabular}{ |c c|c c|c c|c c|c c| }
\hline
N=392 &  & Q1 & Q2 & Q1 & Q2 & Q1 & Q2 & Q1 & Q2 \\ \hline
 \multicolumn{2}{|c|}{ } &
 \multicolumn{2}{|c|}{$E$ }&
 \multicolumn{2}{|c|}{$O\to E$ }&
 \multicolumn{2}{|c|}{$E\to O$ } &
 \multicolumn{2}{|c|}{$O$ }\\
 \hline
Vases & av. & 0.85 & 0.73 & 0.76 & 0.54 & 0.15 & 0.27 & 0.24 & 0.46 \\
 & std. & 0.30 & 0.35 & 0.34 & 0.38 & 0.30 & 0.35 & 0.34 & 0.38 \\
 \hline
 Sofas & av. & 0.87 & 0.47 & 0.76 & 0.48 & 0.13 & 0.53 & 0.24 & 0.52 \\
 & std. & 0.28 & 0.41 & 0.36 & 0.41 & 0.28 & 0.41 & 0.36 & 0.41 \\
  \hline
   Pillows & av. & 0.92 & 0.66 & 0.88 & 0.54 & 0.08 & 0.34 & 0.12 & 0.46 \\
 & std. & 0.20 & 0.39 & 0.24 & 0.42 & 0.20 & 0.39 & 0.24 & 0.42 \\
 \hline
    Lamps & av. & 0.94 & 0.67 & 0.93 & 0.51 & 0.06 & 0.33 & 0.07 & 0.49 \\
 & std. & 0.17 & 0.35 & 0.17 & 0.37 & 0.17 & 0.35 & 0.17 & 0.37 \\ \hline
    All & av. & 0.89 & 0.63 & 0.83 & 0.52 & 0.11 & 0.37 & 0.17 & 0.48 \\
 & std. & 0.13 & 0.22 & 0.16 & 0.23 & 0.13 & 0.22 & 0.16 & 0.23 \\
  \hline
\end{tabular}
  \caption{Questionnaire's third section. For each question Q1 and Q2, the table reports the average score and its standard deviation obtained by objects (vase, sofa, pillow, or lamp, both individually taken and all together) when performing pairwise comparisons $E$ vs. $E \to O$ and $O$ vs. $O \to E$, N=392 respondents. For each question, the sum of the $E$ and $E \to O$ columns (respectively, $O$ and $O \to E$ columns) is one.}
  \label{Table 3}
\end{table}


\begin{table}
\begin{tabular}{ |c c|c c|c c|c c|c c| }
\hline
N=117 &  & Q1 & Q2 & Q1 & Q2 & Q1 & Q2 & Q1 & Q2 \\ \hline
 \multicolumn{2}{|c|}{ } &
 \multicolumn{2}{|c|}{$E$ }&
 \multicolumn{2}{|c|}{$O\to E$ }&
 \multicolumn{2}{|c|}{$E\to O$ } &
 \multicolumn{2}{|c|}{$O$ }\\
 \hline
Vases & av. & 0.84 & 0.73 & 0.75 & 0.56 & 0.16 & 0.27 & 0.25 & 0.44 \\
 & std. & 0.30 & 0.33 & 0.35 & 0.37 & 0.30 & 0.33 & 0.35 & 0.37 \\
 \hline
 Sofas & av. & 0.89 & 0.61 & 0.82 & 0.66 & 0.11 & 0.39 & 0.18 & 0.34 \\
 & std. & 0.26 & 0.40 & 0.32 & 0.39 & 0.26 & 0.40 & 0.32 & 0.39 \\
  \hline
   Pillows & av. & 0.95 & 0.69 & 0.86 & 0.57 & 0.05 & 0.31 & 0.14 & 0.43 \\
 & std. & 0.15 & 0.35 & 0.25 & 0.39 & 0.15 & 0.35 & 0.32 & 0.39 \\
 \hline
    Lamps & av. & 0.96 & 0.69 & 0.96 & 0.51 & 0.04 & 0.31 & 0.04 & 0.49 \\
 & std. & 0.13 & 0.35 & 0.14 & 0.40 & 0.13 & 0.35 & 0.14 & 0.40 \\ \hline
    All & av. & 0.91 & 0.68 & 0.85 & 0.58 & 0.09 & 0.32 & 0.15 & 0.42 \\
 & std. & 0.12 & 0.21 & 0.16 & 0.22 & 0.12 & 0.21 & 0.16 & 0.22 \\
  \hline
\end{tabular}
  \caption{Questionnaire's third Section, age$<30$, N=117. Data are organized as in Table \ref{Table 3}.}
  \label{Table 4}
\end{table}

\subsection{Third Section: Comparing culturized and non-culturized objects}
\label{Comparing culturized and non-culturized objects} 

Table \ref{Table 3} reports the scores of each object class (vase, sofa, pillow, and lamp) and group ($E$, $O \to E$, $E \to O$, $O$) in the third section of the questionnaire, averaged over N respondents. 

\subsubsection{Evaluating Hypothesis 1}
\label{Evaluating Hypothesis 1a}

To search for evidence supporting Hypothesis 1, we considered Q1: `Which object best represents European culture and tradition?' 
Specifically, we showed each respondent two $E$ vs. $E \to O$ and two $O$ vs. $O \to E$ comparisons for each object class and assigned a score of $1$ to the winner of each comparison (i.e., we always compared the original object with its culturized counterpart). 
After averaging over two comparisons and N respondents per object class/group, it can be verified that the maximum score in each cell of the table is 1 (which happens if a group, say $E$, is always preferred to the competing group $E \to O$ by all respondents for that object class), and the sum of two competing groups (say, $E$ and $E \to O$) is 1 as well. 
Analyses were performed using a one-sample T-test: in the $E$ vs. $E \to O$ comparisons, we checked if the average number of times that $E$ was selected $\mu_{E} > 0.5$ with $\alpha=0.05$; in the $O$ vs. $O \to E$ comparison, we checked if $\mu_{O \to E} > 0.5$.


It can be observed that, concerning Q1 (`Which object best represents European culture and tradition?'), 
\begin{itemize}
\item for any object class, $E$ objects were selected, on average, more than half of the time,
\item for any object class, $O \to E$ objects were selected, on average, more than half of the time.
\end{itemize}

The one-tail, one-sample T-test returned that 

\begin{itemize}
\item $\mu_{E} > 0.5$ with $p<0.05$ for any objects individually taken and for all objects taken together, 
\item $\mu_{O \to E} > 0.5$ with $p<0.05$ for any objects individually taken and for all objects taken together,
\end{itemize}
which supports Hypothesis 1 since $E$ and $O\to E$ classes are recognized as more ``European" than their respective counterparts.

\subsubsection{Evaluating Hypothesis 2}
\label{Evaluating Hypothesis 2a}
Concerning Q2 (`Which object do you like the most?'), 
\begin{itemize}
\item for vases, pillows, lamps, $E$ objects were selected, on average, more than half of the time,
\item the same happened for the $O \to E$ group, even if average scores are lower than the $E$ group,
\end{itemize}

The one-tail, one-sample T-test returned that

\begin{itemize}
\item $\mu_{E} > 0.5$ with $p<0.05$ for vases, pillows, lamps, as well as for all objects taken together, 
\item $\mu_{O \to E} > 0.5$ with $p<0.05$ for vases and pillows, $\mu_{O \to E} = 0.5$ with $p<0.05$ for lamps, $\mu_{O \to E} > 0.5$ with $p=0.052$ for all objects together,
\end{itemize}
which supports Hypothesis 2 only partially because, in some cases, the preference for $O \to E$ with respect to $O$ objects is not statistically significant with $p<0.05$. 

\subsubsection{Stratified analysis}
\label{Evaluating Hypothesis stratified a}
We performed a stratified analysis controlling for age: results with age $< 30$ (N=117) are in Table \ref{Table 4}.
All scores totaled by the $E$ and $O \to E$ groups tend to increase. When analyzing Q2 for the $O \to E$ group, it holds $\mu_{O \to E} > 0.5$ with $p<0.05$ for vases, sofas, pillows, and for all objects taken together. However, it holds $\mu_{O \to E} = 0.5$ with $p<0.05$ for lamps. Results almost fully support Hypothesis 2.

\subsubsection{Evaluating Hypothesis 3}
\label{Evaluating Hypothesis 3a}
To search for evidence supporting Hypothesis 3, we considered Q2: `Which object do you like the most?' for the whole sample with N=392. 
 Specifically, as we previously did in section \ref{Recognizing culture and expressing preferences}, we considered the N participants and the four groups ($E$, $O \to E$, $E \to O$, and $O$). For each participant, in this case, we computed only the average score for all objects belonging to the $E$ and $O \to E$ groups: this process yielded two N-sized vectors $A_E$ and $A_{O\to E}$, one per group ($A_{E\to O}$ and $A_{O}$ can simply be computed as the unitary vector 1 - $A_{E}$ and 1 - $A_{O \to E}$, as a consequence of how scores are assigned to pairwise comparisons). As in the previous case, the N elements of a vector represent the average score assigned to that group by the N participants.

Next, it was sufficient to correlate the two N-sized vectors $A_E$ and $A_{O\to E}$ by computing Pearson's $\rho_{E, O \to E}$, since it holds $\rho_{E, E \to O}=-1$ and $\rho_{E, O}=-\rho_{E, O \to E}$.
The analysis returned $\rho_{E, O \to E} = 0.66$ (and, consequently, $\rho_{E, O}=-0.66$) supporting Hypothesis 3: respondents giving high scores to $E$ objects tended to prefer $O \to E$ objects as well.


\begin{table}
\begin{tabular}{ |c c|c c||c c|c c| }
\hline
N=392 &  & Q1 & Q2 & N=117 &  & Q1 & Q2  \\ 
 \hline
 &  & $Gcult$ & $Pcult$ & & & $Gcult$ & $Pcult$ \\
  \hline
 & av. & 0.56 & 0.44 & & av. & 0.64 & 0.36 \\
  & std. & 0.17 & 0.17 & & std. & 0.16 & 0.16 \\
  \hline
\end{tabular}
  \caption{Questionnaire's fourth Section. Left: all age ranges, N=392. Right: age $< 30$, N=117. The table reports the average score and its standard deviation relative to $Gcult$ and $Pcult$ environments, computed over N respondents. The sum of the  $Gcult$ and $Pcult$ columns is one.}
  \label{Table5}
\end{table}

\subsection{Fourth Section: Comparing the realism of environments }
\label{ Comparing realism of environments } 

Table \ref{Table5} reports the scores of each environment ($Gcult$ and $Pcult$) in the fourth section of the questionnaire, averaged over N=392 respondents. 

\subsubsection{Evaluating Hypothesis 4}
\label{Evaluating Hypothesis 4}

To search for evidence supporting Hypothesis 4, we asked Q1 (`Which image looks more realistic?'). Specifically, we showed each respondent 8 $Gcult$ vs. $Pcult$ comparisons and assigned a score of $1$ to the winner of each comparison. After averaging the scores given by each respondent in the eight comparisons and then averaging over N respondents, the 
sum of the $Gcult$ and $Pcult$ columns is 1. 

Analyses were performed using a one-sample T-test: we checked if the average number of times that $Gcult$ environments were selected $\mu_{Gcult} > 0.5$ with $\alpha=0.05$. 
 
It can be observed that:
\begin{itemize}
\item $Gcult$ environments were selected, on average, more than half of the time;
\item the average $Gcult$ score tends to increase when performing a stratified analysis for age; the maximum value corresponded to respondents whose age $< 30$.
\end{itemize}

The one-tail, one-sample T-test returned $\mu_{Gcult} > 0.5$ with $p<.001$ in all cases, fully supporting Hypothesis 4 that culturized environments are perceived as more realistic.

\section{Discussion}
\label{Discussion}

We start with general considerations and then draw some conclusions about each Hypothesis. 

\subsection{General considerations}
\label{General considerations d}
The data show that responses may vary significantly depending on the object class considered and different objects within the same class. 
Generally speaking, and according to expectations, Italian participants recognize $E$ and $O\to E$ objects as European and exhibit a conservative preference for familiar $E$ objects.
However, in some cases, the original, non-European, $O$ lamp is preferred to the culturized $O\to E$ version; in other cases, the opposite is true. This result is particularly evident when objects are not pairwise compared: people express their preferences, notwithstanding their cultural belonging, depending on their tastes, or just because an object looks better than another, thanks to the photographer's skill. The pairwise comparisons in the second section somehow control for confounding variables by presenting two objects with very similar structure, dimensions, perspective, lighting, and sometimes texture and altering only the cultural context. However, based on the results, we cannot exclude that choosing different objects for pairwise comparisons might produce different outcomes, see Remark \ref{R5}

\subsection{Hypothesis 1}
\label{Hypothesis 1 d}
It is evident from responses to Q1 in the second and third questionnaire's sections that $E$ and $O \to E$  objects are rated as belonging to the European culture to a higher degree than their non-European counterpart, which is in line with Hypothesis 1. As stated in Remark \ref{R4}, this is very important because it supports our conjecture that participants share a common understanding of what is $E$ and what is $O$, laying the basis for the following questions and related analyses. Interestingly, $O \to E$ objects achieve a lower average score than $E$ ones: this is likely because GANs preserve some elements of the $O$ source image in the translation and therefore $O \to E$ objects are perceived by respondents as not perfectly matching their expectations for a European object.       

\subsection{Hypothesis 2}
\label{Hypothesis 2 d}
It is evident from the responses to Q2 in the second and third sections that $E$ objects are preferred, on average, to all other objects, which is in line with Hypothesis 2. 
The only cases for which this result is not confirmed are the lamp class in the second section and the sofa class in the third section. 
However, it is impossible to draw similar conclusions when focussing on culturized objects: $O \to E$ vases are always preferred to $O$ vases, but the results vary for other object classes. 
The individual preferences of respondents may explain this result: due to immigration, it is common in Italian shops and houses to see Moroccan, Indian, and Chinese pillows and lamps. Despite their conservative attitude \citep{Rubera2011459}, Italians are getting accustomed to these objects and perceive them as familiar, and many respondents may have a favorable bias toward colorful objects from around the world. A similar explanation may hold when considering the sofa class: the classic style sofas shown in the questionnaire are not very common in ordinary Italian houses where a more sober design usually prevails. 
Also, it is interesting that the strongest bias towards $O \to E$ objects is observed in the vase class. Greek vases are correctly perceived as European artifacts, and, on average, respondents prefer them to Far-eastern vases. This result is in line with the explanation above. According to the authors' experience, decorated Far-eastern vases, as shown in Figure \ref{all}, are less frequently encountered in ordinary Italian houses than Moroccan and Indian pillows and lamps. Then, when comparing Greek and Far-eastern vases, the choice might be more determined by cultural belonging than individual preferences and familiarity.  

In the stratified analysis in the second and third questionnaire sections, younger respondents $<30$ tend to prefer $E$ and $O \to E$ objects. This result can have multiple explanations. Among others, this more conservative attitude might be explained by the lesser experience with other cultures and the limited interest younger people may have in furniture items. Many of them have never faced the problem of furnishing a house, possibly because they still live with their parents -- in Italy, the average age when people leave their parents' house is around 30. This conjecture is also coherent with recent studies about openness, see \citep{Shwaba17} and the references therein, arguing that exploration
experiences that tend to put people into novel situations are associated with increased openness (studying abroad in college and attending cultural activities are two prominent examples).

\subsection{Hypothesis 3}
\label{Hypothesis 3 d}
We found a positive correlation between respondents giving a high score to $E$ objects and $O \to E$ objects, which confirms Hypothesis 3. Even if drawing strong conclusions about the preferences of Italians is not possible (and not desirable as it might lead to stereotypes), this result confirms the possibility of identifying people that prefer culturized images that match their background during interactions with robots. In short, a person who prefers European objects tends to favor both $E$ and $O \to E$, which motivates the culturization process.

\subsection{Hypothesis 4}
\label{Hypothesis 4 d}
It is evident from the responses to Q1 in the fourth section that environments modified by segmenting objects, modifying them with GANs, and re-inserting them into the original image are perceived as more realistic, which confirms Hypothesis 4. Some respondents motivated their choice with a written comment complaining that objects in images not modified with GANs have a weird perspective and do not merge well with the background. Not surprisingly, younger people $< 30$ appear more skilled in distinguishing between GAN-modified and non-GAN-modified objects and environments, with a more marked preference for the former. 

\section{Conclusion}
\label{sec:conclusion}

This article introduced the concept of image ``culturization," proposed a process for object culturization based on GANs, and posed the research question of whether people prefer images that were culturized to make them coherent with their culture.
We preliminary explored this research question by 
analyzing the preferences of Italian participants towards objects belonging to European and non-European cultures and the realism of culturized environments. Overall, experiments motivate our intention to proceed further along this path: even if, as expected, not all participants prefer objects belonging to their cultural background, those who prefer European over non-European objects also tend to have a positive attitude towards objects that we culturized to be perceived as European.
    
We already discussed the main limitations of our work, which are summarized here.

First, we focussed on the culturization of individual objects instead of complete scenes. When considering complex environments, we implicitly assumed that it is possible to culturize an environment by modifying the objects that it contains. This conjecture needs to be revised in the future: the color, material, pattern, and form of walls and other architectural features may also play a role. However, given the GAN technology at the time this research was done, this was the best we could do. Recent AI art generators\footnote{https://www.demandsage.com/ai-image-generators/} might evolve the capability to add cultural brushstrokes to complex photorealistic environments. 

Second, we performed experiments with Italian participants only. As discussed throughout the article, there are unique aspects of Italian culture that make it an interesting and relevant case study for exploring the impact of image culturization. However, additional research will be needed to ensure the transferability of findings to other cultures.

Third, when preparing the questionnaires, we selected a subset of objects and environments that seemed promising to produce different reactions in participants. However, we did not validate our questionnaire to capture a hypothetic construct ``positive attitude towards European object". Then, we cannot ensure that results may be generalized to any set of objects, even if the pairwise comparisons showing the original and culturized versions of the same object may somehow control confounding variables. 

Finally, from a technical standpoint, the culturization process is not entirely automated since culturized objects need to be manually reinserted into the environment. However, we do not expect this to impact the hypotheses tested with experiments.

Future work will address the aforementioned issues, and 
expand our research to real-time human-robot interaction with culturized images. 

\section*{Statements and Declarations}
\label{sec:statements}

\subsection*{Ethics} 
The research involves the use of an anonymous online survey. The information obtained is recorded in such a manner that the identity of the human subjects cannot readily be ascertained, directly or through identifiers linked to the subjects.


\subsection*{Relation to prior publications} 
This article has no relation to the authors' prior publications and has not been previously submitted to a conference or a journal.

\subsection*{Conflict of interest} 
The authors declare that they have no conflict of interest.

\subsection*{Data availability} 
The datasets generated during and/or analysed during the current study are available from the corresponding author on reasonable request.

\subsection*{Contribution} 
Conceptualization: C.T. Recchiuto, A. Sgorbissa; Methodology: G. Zaino, C.T. Recchiuto, A. Sgorbissa;  Software: G. Zaino; Investigation: G. Zaino, A. Sgorbissa; Writing: G. Zaino, A. Sgorbissa; Supervision: A. Sgorbissa.

\subsection*{Funding} 
This research did not receive any specific grant from funding agencies in the public, commercial, or
not-for-profit sectors.



\begin{thebibliography}{97}
\ifx \bisbn   \undefined \def \bisbn  #1{ISBN #1}\fi
\ifx \binits  \undefined \def \binits#1{#1}\fi
\ifx \bauthor  \undefined \def \bauthor#1{#1}\fi
\ifx \batitle  \undefined \def \batitle#1{#1}\fi
\ifx \bjtitle  \undefined \def \bjtitle#1{#1}\fi
\ifx \bvolume  \undefined \def \bvolume#1{\textbf{#1}}\fi
\ifx \byear  \undefined \def \byear#1{#1}\fi
\ifx \bissue  \undefined \def \bissue#1{#1}\fi
\ifx \bfpage  \undefined \def \bfpage#1{#1}\fi
\ifx \blpage  \undefined \def \blpage #1{#1}\fi
\ifx \burl  \undefined \def \burl#1{\textsf{#1}}\fi
\ifx \doiurl  \undefined \def \doiurl#1{\url{https://doi.org/#1}}\fi
\ifx \betal  \undefined \def \betal{\textit{et al.}}\fi
\ifx \binstitute  \undefined \def \binstitute#1{#1}\fi
\ifx \binstitutionaled  \undefined \def \binstitutionaled#1{#1}\fi
\ifx \bctitle  \undefined \def \bctitle#1{#1}\fi
\ifx \beditor  \undefined \def \beditor#1{#1}\fi
\ifx \bpublisher  \undefined \def \bpublisher#1{#1}\fi
\ifx \bbtitle  \undefined \def \bbtitle#1{#1}\fi
\ifx \bedition  \undefined \def \bedition#1{#1}\fi
\ifx \bseriesno  \undefined \def \bseriesno#1{#1}\fi
\ifx \blocation  \undefined \def \blocation#1{#1}\fi
\ifx \bsertitle  \undefined \def \bsertitle#1{#1}\fi
\ifx \bsnm \undefined \def \bsnm#1{#1}\fi
\ifx \bsuffix \undefined \def \bsuffix#1{#1}\fi
\ifx \bparticle \undefined \def \bparticle#1{#1}\fi
\ifx \barticle \undefined \def \barticle#1{#1}\fi
\bibcommenthead
\ifx \bconfdate \undefined \def \bconfdate #1{#1}\fi
\ifx \botherref \undefined \def \botherref #1{#1}\fi
\ifx \url \undefined \def \url#1{\textsf{#1}}\fi
\ifx \bchapter \undefined \def \bchapter#1{#1}\fi
\ifx \bbook \undefined \def \bbook#1{#1}\fi
\ifx \bcomment \undefined \def \bcomment#1{#1}\fi
\ifx \oauthor \undefined \def \oauthor#1{#1}\fi
\ifx \citeauthoryear \undefined \def \citeauthoryear#1{#1}\fi
\ifx \endbibitem  \undefined \def \endbibitem {}\fi
\ifx \bconflocation  \undefined \def \bconflocation#1{#1}\fi
\ifx \arxivurl  \undefined \def \arxivurl#1{\textsf{#1}}\fi
\csname PreBibitemsHook\endcsname

\bibitem[\protect\citeauthoryear{Geertz}{1973}]{Geertz73}
\begin{bbook}
\bauthor{\bsnm{Geertz}, \binits{C.}}:
\bbtitle{The Interpretation of Cultures, 2nd Ed.}
\bpublisher{New York},
\blocation{Basic Books}
(\byear{1973})
\end{bbook}
\endbibitem

\bibitem[\protect\citeauthoryear{Hofstede}{1980}]{Hofstede80}
\begin{bbook}
\bauthor{\bsnm{Hofstede}, \binits{G.}}:
\bbtitle{Culture's Consequences: International Differences in Work-related
  Values}.
\bpublisher{Beverly Hills, CA},
\blocation{Sage}
(\byear{1980})
\end{bbook}
\endbibitem

\bibitem[\protect\citeauthoryear{Leininger}{1988}]{Leininger1988}
\begin{barticle}
\bauthor{\bsnm{Leininger}, \binits{M.M.}}:
\batitle{Leininger's theory of nursing: Cultural care diversity and
  universality}.
\bjtitle{Nurs. Sci. Q.}
\bvolume{1}(\bissue{4}),
\bfpage{152}--\blpage{160}
(\byear{1988})
\doiurl{10.1177/089431848800100408}
\end{barticle}
\endbibitem

\bibitem[\protect\citeauthoryear{Schwartz}{1992}]{Schwartz19921}
\begin{barticle}
\bauthor{\bsnm{Schwartz}, \binits{S.H.}}:
\batitle{Universals in the content and structure of values: Theoretical
  advances and empirical tests in 20 countries}.
\bjtitle{Adv. Exp. Soc. Psychol.}
\bvolume{25}(\bissue{C}),
\bfpage{1}--\blpage{65}
(\byear{1992})
\doiurl{10.1016/S0065-2601(08)60281-6} .
\bcomment{cited By 8894}
\end{barticle}
\endbibitem

\bibitem[\protect\citeauthoryear{Henare et~al.}{2007}]{Henare2007}
\begin{bbook}
\bauthor{\bsnm{Henare}, \binits{A.}},
\bauthor{\bsnm{Holbraad}, \binits{M.}},
\bauthor{\bsnm{Wastell}, \binits{S.}}:
\bbtitle{Thinking Through Things: Theorising Artefacts Ethnographically}.
\bpublisher{London},
\blocation{Routledge}
(\byear{2007})
\end{bbook}
\endbibitem

\bibitem[\protect\citeauthoryear{Betancourt et~al.}{2003}]{Betancourt2003}
\begin{barticle}
\bauthor{\bsnm{Betancourt}, \binits{J.R.}},
\bauthor{\bsnm{Green}, \binits{A.R.}},
\bauthor{\bsnm{Carrillo}, \binits{J.E.}},
\bauthor{\bsnm{Ananeh-Firempong~II}, \binits{O.}}:
\batitle{Defining cultural competence: A practical framework for addressing
  racial/ethnic disparities in health and health care}.
\bjtitle{Public Health Rep.}
\bvolume{118}(\bissue{4}),
\bfpage{293}--\blpage{302}
(\byear{2003})
\doiurl{10.1093/phr/118.4.293}
\end{barticle}
\endbibitem

\bibitem[\protect\citeauthoryear{Bruno et~al.}{2017}]{Bruno2017}
\begin{bchapter}
\bauthor{\bsnm{Bruno}, \binits{B.}},
\bauthor{\bsnm{Chong}, \binits{N.Y.}},
\bauthor{\bsnm{Kamide}, \binits{H.}},
\bauthor{\bsnm{Kanoria}, \binits{S.}},
\bauthor{\bsnm{Lee}, \binits{J.}},
\bauthor{\bsnm{Lim}, \binits{Y.}},
\bauthor{\bsnm{Pandey}, \binits{A.K.}},
\bauthor{\bsnm{Papadopoulos}, \binits{C.}},
\bauthor{\bsnm{Papadopoulos}, \binits{I.}},
\bauthor{\bsnm{Pecora}, \binits{F.}},
\bauthor{\bsnm{Saffiotti}, \binits{A.}},
\bauthor{\bsnm{Sgorbissa}, \binits{A.}}:
\bctitle{Paving the way for culturally competent robots: A position paper}.
In: \bbtitle{Proc. 17th IEEE Int. Symp. on Robot and Human Interactive
  Communication, RO-MAN'17},
\bconflocation{Lisbon, Portugal},
pp. \bfpage{553}--\blpage{560}
(\byear{2017}).
\doiurl{10.1109/ROMAN.2017.8172357}
\end{bchapter}
\endbibitem

\bibitem[\protect\citeauthoryear{Bruno et~al.}{2019}]{Bruno2019}
\begin{barticle}
\bauthor{\bsnm{Bruno}, \binits{B.}},
\bauthor{\bsnm{Recchiuto}, \binits{C.T.}},
\bauthor{\bsnm{Papadopoulos}, \binits{I.}},
\bauthor{\bsnm{Saffiotti}, \binits{A.}},
\bauthor{\bsnm{Koulouglioti}, \binits{C.}},
\bauthor{\bsnm{Menicatti}, \binits{R.}},
\bauthor{\bsnm{Mastrogiovanni}, \binits{F.}},
\bauthor{\bsnm{Zaccaria}, \binits{R.}},
\bauthor{\bsnm{Sgorbissa}, \binits{A.}}:
\batitle{Knowledge representation for culturally competent personal robots:
  Requirements, design principles, implementation, and assessment}.
\bjtitle{Int. J. Soc. Robot}
\bvolume{11}(\bissue{3}),
\bfpage{515}--\blpage{538}
(\byear{2019})
\doiurl{10.1007/s12369-019-00519-w}
\end{barticle}
\endbibitem

\bibitem[\protect\citeauthoryear{Knappett}{2005}]{Knappett2005}
\begin{bbook}
\bauthor{\bsnm{Knappett}, \binits{C.}}:
\bbtitle{Thinking Through Material Culture: An Interdisciplinary Perspective},
pp. \bfpage{1}--\blpage{202}.
\bpublisher{University of Pennsylvania Press},
\blocation{Pennsylvania, USA}
(\byear{2005})
\end{bbook}
\endbibitem

\bibitem[\protect\citeauthoryear{Robb}{2015}]{Robb2015}
\begin{barticle}
\bauthor{\bsnm{Robb}, \binits{J.}}:
\batitle{What do things want? object design as a middle range theory of
  material culture}.
\bjtitle{Archeol. Pap. Am. Anthropol. Assoc.}
\bvolume{26}(\bissue{1}),
\bfpage{166}--\blpage{180}
(\byear{2015})
\doiurl{10.1111/APAA.12069}
\end{barticle}
\endbibitem

\bibitem[\protect\citeauthoryear{Hekkert and Leder}{2008}]{Hekkert2008}
\begin{bchapter}
\bauthor{\bsnm{Hekkert}, \binits{P.}},
\bauthor{\bsnm{Leder}, \binits{H.}}:
\bctitle{Product aesthetics}.
In: \beditor{\bsnm{Schifferstein}, \binits{H.N.J.}},
\beditor{\bsnm{Hekkert}, \binits{P.}} (eds.)
\bbtitle{Product Experience},
pp. \bfpage{259}--\blpage{285}.
\bpublisher{Elsevier},
\blocation{San Diego}
(\byear{2008}).
\doiurl{10.1016/B978-008045089-6.50013-7}
\end{bchapter}
\endbibitem

\bibitem[\protect\citeauthoryear{Hsu et~al.}{2011}]{Hsu2011}
\begin{barticle}
\bauthor{\bsnm{Hsu}, \binits{C.-H.}},
\bauthor{\bsnm{Lin}, \binits{C.-L.}},
\bauthor{\bsnm{Lin}, \binits{R.}}:
\batitle{A study of framework and process development for cultural product
  design}.
\bjtitle{Lecture Notes in Computer Science}
\bvolume{6775 LNCS},
\bfpage{55}--\blpage{64}
(\byear{2011})
\end{barticle}
\endbibitem

\bibitem[\protect\citeauthoryear{Chai et~al.}{2015}]{Chai2015}
\begin{barticle}
\bauthor{\bsnm{Chai}, \binits{C.}},
\bauthor{\bsnm{Bao}, \binits{D.}},
\bauthor{\bsnm{Sun}, \binits{L.}},
\bauthor{\bsnm{Cao}, \binits{Y.}}:
\batitle{The relative effects of different dimensions of traditional cultural
  elements on customer product satisfaction}.
\bjtitle{Int. J. Ind. Ergon.}
\bvolume{48},
\bfpage{77}--\blpage{88}
(\byear{2015})
\doiurl{10.1016/j.ergon.2015.04.001}
\end{barticle}
\endbibitem

\bibitem[\protect\citeauthoryear{Goodfellow et~al.}{2014}]{Goodfellow2014}
\begin{bchapter}
\bauthor{\bsnm{Goodfellow}, \binits{I.J.}},
\bauthor{\bsnm{Pouget-Abadie}, \binits{J.}},
\bauthor{\bsnm{Mirza}, \binits{M.}},
\bauthor{\bsnm{Xu}, \binits{B.}},
\bauthor{\bsnm{Warde-Farley}, \binits{D.}},
\bauthor{\bsnm{Ozair}, \binits{S.}},
\bauthor{\bsnm{Courville}, \binits{A.}},
\bauthor{\bsnm{Bengio}, \binits{Y.}}:
\bctitle{{Generative Adversarial Networks}}.
In: \bbtitle{Proc. 2014 Conf. on Neural Information Processing Systems,
  NIPS'14},
\bconflocation{Montreal, Canada},
pp. \bfpage{2672}--\blpage{2680}
(\byear{2014}).
\doiurl{10.48550/arXiv.1406.2661}
\end{bchapter}
\endbibitem

\bibitem[\protect\citeauthoryear{Isola et~al.}{2017}]{Isola2017}
\begin{bchapter}
\bauthor{\bsnm{Isola}, \binits{P.}},
\bauthor{\bsnm{Zhu}, \binits{J.}},
\bauthor{\bsnm{Zhou}, \binits{T.}},
\bauthor{\bsnm{Efros}, \binits{A.A.}}:
\bctitle{Image-to-image translation with conditional adversarial networks}.
In: \bbtitle{Proc. 2017 IEEE Conf. on Computer Vision and Pattern Recognition,
  CVPR'17},
\bconflocation{Honolulu, HI, USA}
(\byear{2017}).
\doiurl{10.1109/CVPR.2017.632}
\end{bchapter}
\endbibitem

\bibitem[\protect\citeauthoryear{Zhu et~al.}{2017}]{Jun-Yan2017}
\begin{bchapter}
\bauthor{\bsnm{Zhu}, \binits{J.}},
\bauthor{\bsnm{Park}, \binits{T.}},
\bauthor{\bsnm{Isola}, \binits{P.}},
\bauthor{\bsnm{Efros}, \binits{A.A.}}:
\bctitle{Unpaired image-to-image translation using cycle-consistent adversarial
  networks}.
In: \bbtitle{Proc. 2017 Int. Conf. on Computer Vision, {ICCV}'17},
\bconflocation{Venice, Italy},
pp. \bfpage{2242}--\blpage{2251}
(\byear{2017}).
\doiurl{10.1109/ICCV.2017.244}
\end{bchapter}
\endbibitem

\bibitem[\protect\citeauthoryear{Kim et~al.}{2017}]{Taeksoo2017}
\begin{bchapter}
\bauthor{\bsnm{Kim}, \binits{T.}},
\bauthor{\bsnm{Cha}, \binits{M.}},
\bauthor{\bsnm{Kim}, \binits{H.}},
\bauthor{\bsnm{Lee}, \binits{J.K.}},
\bauthor{\bsnm{Kim}, \binits{J.}}:
\bctitle{Learning to discover cross-domain relations with generative
  adversarial networks}.
In: \bbtitle{Proc. 2017 Int. Conf. on Machine Learning, ICML'17},
vol. \bseriesno{70}.
\bconflocation{Sydney, Australia}
(\byear{2017}).
\doiurl{10.48550/arXiv.1703.05192}
\end{bchapter}
\endbibitem

\bibitem[\protect\citeauthoryear{Mejjati et~al.}{2018}]{Mejjati2018}
\begin{bchapter}
\bauthor{\bsnm{Mejjati}, \binits{Y.A.}},
\bauthor{\bsnm{Richardt}, \binits{C.}},
\bauthor{\bsnm{Cosker}, \binits{D.}},
\bauthor{\bsnm{Tompkin}, \binits{J.}},
\bauthor{\bsnm{Kim}, \binits{K.I.}}:
\bctitle{Unsupervised attention-guided image-to-image translation}.
In: \bbtitle{Proc. 2018 Conf. on Neural Information Processing Systems,
  NIPS'18},
\bconflocation{Montréal, Canada},
pp. \bfpage{3693}--\blpage{3703}
(\byear{2018}).
\doiurl{10.48550/arXiv.1806.02311}
\end{bchapter}
\endbibitem

\bibitem[\protect\citeauthoryear{Mo et~al.}{2019}]{Sangwoo2018}
\begin{bchapter}
\bauthor{\bsnm{Mo}, \binits{S.}},
\bauthor{\bsnm{Cho}, \binits{M.}},
\bauthor{\bsnm{Shin}, \binits{J.}}:
\bctitle{{InstaGAN}: Instance-aware image-to-image translation}.
In: \bbtitle{Proc. 2019 Int. Conf. on Learning Representations, ICLR'19},
\bconflocation{New Orleans, LA, USA}
(\byear{2019}).
\doiurl{10.48550/arXiv.1812.10889}
\end{bchapter}
\endbibitem

\bibitem[\protect\citeauthoryear{Liu et~al.}{2017}]{Unit2017}
\begin{bchapter}
\bauthor{\bsnm{Liu}, \binits{M.}},
\bauthor{\bsnm{Breuel}, \binits{T.}},
\bauthor{\bsnm{Kautz}, \binits{J.}}:
\bctitle{Unsupervised image-to-image translation networks}.
In: \bbtitle{Proc. 2017 Conf. on Neural Information Processing Systems,
  NIPS'17},
\bconflocation{Long Beach, CA, USA},
pp. \bfpage{700}--\blpage{708}
(\byear{2017}).
\doiurl{10.48550/arXiv.1703.00848}
\end{bchapter}
\endbibitem

\bibitem[\protect\citeauthoryear{Rubera et~al.}{2011}]{Rubera2011459}
\begin{barticle}
\bauthor{\bsnm{Rubera}, \binits{G.}},
\bauthor{\bsnm{Ordanini}, \binits{A.}},
\bauthor{\bsnm{Griffith}, \binits{D.A.}}:
\batitle{Incorporating cultural values for understanding the influence of
  perceived product creativity on intention to buy: An examination in italy and
  the us}.
\bjtitle{J. Int. Bus.}
\bvolume{42}(\bissue{4}),
\bfpage{459}--\blpage{476}
(\byear{2011})
\doiurl{10.1057/jibs.2011.3} .
\bcomment{cited By 53}
\end{barticle}
\endbibitem

\bibitem[\protect\citeauthoryear{Pei-Luen and Rau}{2022}]{CCD2022}
\begin{bbook}
\beditor{\bsnm{Pei-Luen}},
\beditor{\bsnm{Rau}, \binits{P.}} (eds.):
\bbtitle{14th Int. Conf. on Cross-Cultural Design, {CCD 2022} Held as Part of
  the 24th {HCI} Int. Conf., {HCI'22}}
vol. \bseriesno{13311 LNCS},
(\byear{2022})
\end{bbook}
\endbibitem

\bibitem[\protect\citeauthoryear{Lin et~al.}{2007}]{Lin2007146}
\begin{bchapter}
\bauthor{\bsnm{Lin}, \binits{R.}},
\bauthor{\bsnm{Sun}, \binits{M.-X.}},
\bauthor{\bsnm{Chang}, \binits{Y.-P.}},
\bauthor{\bsnm{Chan}, \binits{Y.-C.}},
\bauthor{\bsnm{Hsieh}, \binits{Y.-C.}},
\bauthor{\bsnm{Huang}, \binits{Y.-C.}}:
\bctitle{Designing ``culture" into modern product: A case study of cultural
  product design}.
In: \bbtitle{Lecture Notes in Computer Science},
vol. \bseriesno{4559 LNCS},
pp. \bfpage{146}--\blpage{153}
(\byear{2007}).
\doiurl{10.1007/978-3-540-73287-7_19}
\end{bchapter}
\endbibitem

\bibitem[\protect\citeauthoryear{Chavan et~al.}{2009}]{Chavan200926}
\begin{barticle}
\bauthor{\bsnm{Chavan}, \binits{A.L.}},
\bauthor{\bsnm{Gorney}, \binits{D.}},
\bauthor{\bsnm{Prabhu}, \binits{B.}},
\bauthor{\bsnm{Arora}, \binits{S.}}:
\batitle{The washing machine that ate my sari - mistakes in cross-cultural
  design}.
\bjtitle{Interactions}
\bvolume{16}(\bissue{1}),
\bfpage{26}--\blpage{31}
(\byear{2009})
\doiurl{10.1145/1456202.1456209}
\end{barticle}
\endbibitem

\bibitem[\protect\citeauthoryear{Winschiers-Theophilus}{2009}]{Winschiers-Theophilus2009665}
\begin{bchapter}
\bauthor{\bsnm{Winschiers-Theophilus}, \binits{H.}}:
\bctitle{The art of cross-cultural design for usability}.
In: \bbtitle{Lecture Notes in Computer Science},
vol. \bseriesno{5614 LNCS},
pp. \bfpage{665}--\blpage{671}
(\byear{2009}).
\doiurl{10.1007/978-3-642-02707-9_75}
\end{bchapter}
\endbibitem

\bibitem[\protect\citeauthoryear{Patrick~Rau et~al.}{2012}]{PatrickRau20121}
\begin{bbook}
\bauthor{\bsnm{Patrick~Rau}, \binits{P.-L.}},
\bauthor{\bsnm{Plocher}, \binits{T.}},
\bauthor{\bsnm{Choong}, \binits{Y.-Y.}}:
\bbtitle{Cross-Cultural Design for IT Products and Services},
pp. \bfpage{1}--\blpage{239}.
\bpublisher{CRC Press}, \blocation{???}
(\byear{2012}).
\doiurl{10.1201/b12679}
\end{bbook}
\endbibitem

\bibitem[\protect\citeauthoryear{Sun}{2012}]{Sun20121}
\begin{bbook}
\bauthor{\bsnm{Sun}, \binits{H.}}:
\bbtitle{Cross-Cultural Technology Design: Creating Culture-Sensitive
  Technology for Local Users},
pp. \bfpage{1}--\blpage{352}.
\bpublisher{Oxford University Pres}, \blocation{???}
(\byear{2012}).
\doiurl{10.1093/acprof:oso/9780199744763.001.0001}
\end{bbook}
\endbibitem

\bibitem[\protect\citeauthoryear{Lightner et~al.}{2002}]{Lightner2002373}
\begin{barticle}
\bauthor{\bsnm{Lightner}, \binits{N.J.}},
\bauthor{\bsnm{Yenisey}, \binits{M.M.}},
\bauthor{\bsnm{Ozok}, \binits{A.A.}},
\bauthor{\bsnm{Salvendy}, \binits{G.}}:
\batitle{Shopping behaviour and preferences in e-commerce of turkish and
  american university students: Implications from cross-cultural design}.
\bjtitle{Behav. Inf. Technol.}
\bvolume{21}(\bissue{6}),
\bfpage{373}--\blpage{385}
(\byear{2002})
\doiurl{10.1080/0144929021000071316}
\end{barticle}
\endbibitem

\bibitem[\protect\citeauthoryear{Santoso et~al.}{2018}]{Santoso2018}
\begin{bchapter}
\bauthor{\bsnm{Santoso}, \binits{H.B.}},
\bauthor{\bsnm{Rachmadiani}, \binits{D.}},
\bauthor{\bsnm{Nurrohmah}, \binits{I.}}:
\bctitle{Cultural consideration for designing e-commerce site interface}.
In: \bbtitle{Proc. 1st Int. Conf. on Computer Applications and Information
  Security, ICCAIS 2018},
\bconflocation{Riyadh, Saudi Arabia}
(\byear{2018}).
\doiurl{10.1109/CAIS.2018.8442039}
\end{bchapter}
\endbibitem

\bibitem[\protect\citeauthoryear{Bonenberg}{2016}]{Bonenberg2016105}
\begin{barticle}
\bauthor{\bsnm{Bonenberg}, \binits{W.}}:
\batitle{Cross-cultural design and its application in architecture}.
\bjtitle{Adv. Intell. Syst. Comput.}
\bvolume{493},
\bfpage{105}--\blpage{114}
(\byear{2016})
\doiurl{10.1007/978-3-319-41941-1_10}
\end{barticle}
\endbibitem

\bibitem[\protect\citeauthoryear{O’Rourke et~al.}{2022}]{ORourke202289}
\begin{barticle}
\bauthor{\bsnm{O’Rourke}, \binits{T.}},
\bauthor{\bsnm{Nash}, \binits{D.}},
\bauthor{\bsnm{Haynes}, \binits{M.}},
\bauthor{\bsnm{Burgess}, \binits{M.}},
\bauthor{\bsnm{Memmott}, \binits{P.}}:
\batitle{Cross-cultural design and healthcare waiting rooms for indigenous
  people in regional australia}.
\bjtitle{Environ. Behav.}
\bvolume{54}(\bissue{1}),
\bfpage{89}--\blpage{115}
(\byear{2022})
\doiurl{10.1177/0013916520952443}
\end{barticle}
\endbibitem

\bibitem[\protect\citeauthoryear{Pan and Yu}{2020}]{Pan2020308}
\begin{barticle}
\bauthor{\bsnm{Pan}, \binits{Y.}},
\bauthor{\bsnm{Yu}, \binits{W.}}:
\batitle{Discussion on aesthetic design in chinese painting based on
  cross-cultural design}.
\bjtitle{Commun. Comput. Inf. Sci.}
\bvolume{1226 CCIS},
\bfpage{308}--\blpage{316}
(\byear{2020})
\doiurl{10.1007/978-3-030-50732-9_41}
\end{barticle}
\endbibitem

\bibitem[\protect\citeauthoryear{Zhou et~al.}{2021}]{Zhou2021373}
\begin{bchapter}
\bauthor{\bsnm{Zhou}, \binits{D.T.}},
\bauthor{\bsnm{Yuan}, \binits{X.F.}},
\bauthor{\bsnm{Wu}, \binits{Y.}},
\bauthor{\bsnm{Pan}, \binits{C.X.}}:
\bctitle{Transforming chinese ancient ship art and cultural features into
  modern yacht: A design dna model of yacht localization design}.
In: \bbtitle{Proc. 5th Int. Conf. on Maritime Technology and Engineering,
  MARTECH 2020},
vol. \bseriesno{1}.
\bconflocation{Lisbon, Portugal},
pp. \bfpage{373}--\blpage{380}
(\byear{2021}).
\doiurl{10.1201/9781003216582-42}
\end{bchapter}
\endbibitem

\bibitem[\protect\citeauthoryear{Asino and Giacumo}{2019}]{Asino2019395}
\begin{bchapter}
\bauthor{\bsnm{Asino}, \binits{T.I.}},
\bauthor{\bsnm{Giacumo}, \binits{L.}}:
\bctitle{Culture and global workplace learning: Foundations of cross-cultural
  design theories and models}.
\bbtitle{The Wiley Handbook of Global Workplace Learning},
pp. \bfpage{395}--\blpage{412}.
\bpublisher{Wiley}, \blocation{???}
(\byear{2019}).
\doiurl{10.1002/9781119227793.ch22}
\end{bchapter}
\endbibitem

\bibitem[\protect\citeauthoryear{Cao et~al.}{2021}]{Cao2021197}
\begin{bchapter}
\bauthor{\bsnm{Cao}, \binits{X.}},
\bauthor{\bsnm{Hsu}, \binits{Y.}},
\bauthor{\bsnm{Wu}, \binits{W.}}:
\bctitle{Cross-cultural design: A set of design heuristics for concept
  generation of sustainable packagings}.
In: \bbtitle{Lecture Notes in Computer Science},
vol. \bseriesno{12771 LNCS},
pp. \bfpage{197}--\blpage{209}
(\byear{2021}).
\doiurl{10.1007/978-3-030-77074-7_16}
\end{bchapter}
\endbibitem

\bibitem[\protect\citeauthoryear{Li and Hölttä-Otto}{2020}]{Li2020779}
\begin{barticle}
\bauthor{\bsnm{Li}, \binits{J.}},
\bauthor{\bsnm{Hölttä-Otto}, \binits{K.}}:
\batitle{The influence of designers’ cultural differences on the empathic
  accuracy of user understanding}.
\bjtitle{Des. J.}
\bvolume{23}(\bissue{5}),
\bfpage{779}--\blpage{796}
(\byear{2020})
\doiurl{10.1080/14606925.2020.1810414}
\end{barticle}
\endbibitem

\bibitem[\protect\citeauthoryear{Wang et~al.}{2020}]{Wang2020233}
\begin{barticle}
\bauthor{\bsnm{Wang}, \binits{W.}},
\bauthor{\bsnm{Bryan-Kinns}, \binits{N.}},
\bauthor{\bsnm{Sheridan}, \binits{J.G.}}:
\batitle{On the role of in-situ making and evaluation in designing across
  cultures}.
\bjtitle{CoDesign}
\bvolume{16}(\bissue{3}),
\bfpage{233}--\blpage{250}
(\byear{2020})
\doiurl{10.1080/15710882.2019.1580296}
\end{barticle}
\endbibitem

\bibitem[\protect\citeauthoryear{Guo et~al.}{2022}]{Guo202220}
\begin{bchapter}
\bauthor{\bsnm{Guo}, \binits{Z.}},
\bauthor{\bsnm{Rau}, \binits{P.-L.P.}},
\bauthor{\bsnm{Heimgärtner}, \binits{R.}}:
\bctitle{The “onion model of human factors”: A theoretical framework for
  cross-cultural design}.
In: \bbtitle{Lecture Notes in Computer Science},
vol. \bseriesno{13324 LNCS},
pp. \bfpage{20}--\blpage{33}
(\byear{2022}).
\doiurl{10.1007/978-3-031-05434-1_2}
\end{bchapter}
\endbibitem

\bibitem[\protect\citeauthoryear{Meier et~al.}{2014}]{Meier2014211}
\begin{bchapter}
\bauthor{\bsnm{Meier}, \binits{A.}},
\bauthor{\bsnm{Goto}, \binits{K.}},
\bauthor{\bsnm{Wörmann}, \binits{M.}}:
\bctitle{Thumbs up to gesture controls? a cross-cultural study on spontaneous
  gestures}.
In: \bbtitle{Lecture Notes in Computer Science},
vol. \bseriesno{8528 LNCS},
pp. \bfpage{211}--\blpage{217}
(\byear{2014}).
\doiurl{10.1007/978-3-319-07308-8_21}
\end{bchapter}
\endbibitem

\bibitem[\protect\citeauthoryear{Jane et~al.}{2017}]{Jane20176794}
\begin{bchapter}
\bauthor{\bsnm{Jane}, \binits{L.E.}},
\bauthor{\bsnm{Ilene}, \binits{L.E.}},
\bauthor{\bsnm{Landay}, \binits{J.A.}},
\bauthor{\bsnm{Cauchard}, \binits{J.R.}}:
\bctitle{Drone \& {Wo}: Cultural influences on human-drone interaction
  techniques}.
In: \bbtitle{Proc. 2017 Conf. on Human Factors in Computing Systems, CHI'17},
vol. \bseriesno{2017-May}.
\bconflocation{Denver, CO, USA},
pp. \bfpage{6794}--\blpage{6799}
(\byear{2017}).
\doiurl{10.1145/3025453.3025755}
\end{bchapter}
\endbibitem

\bibitem[\protect\citeauthoryear{Urakami}{2019}]{Urakami2019615}
\begin{barticle}
\bauthor{\bsnm{Urakami}, \binits{J.}}:
\batitle{Towards cross-cultural design of interfaces: Preferences in interface
  design between japanese and european users}.
\bjtitle{Adv. Intell. Syst. Comput.}
\bvolume{783},
\bfpage{615}--\blpage{625}
(\byear{2019})
\doiurl{10.1007/978-3-319-94709-9_60}
\end{barticle}
\endbibitem

\bibitem[\protect\citeauthoryear{Young et~al.}{2012}]{Young2012564}
\begin{barticle}
\bauthor{\bsnm{Young}, \binits{K.L.}},
\bauthor{\bsnm{Rudin-Brown}, \binits{C.M.}},
\bauthor{\bsnm{Lenné}, \binits{M.G.}},
\bauthor{\bsnm{Williamson}, \binits{A.R.}}:
\batitle{The implications of cross-regional differences for the design of
  in-vehicle information systems: A comparison of australian and chinese
  drivers}.
\bjtitle{Appl. Ergon.}
\bvolume{43}(\bissue{3}),
\bfpage{564}--\blpage{573}
(\byear{2012})
\doiurl{10.1016/j.apergo.2011.09.001}
\end{barticle}
\endbibitem

\bibitem[\protect\citeauthoryear{Koratpallikar and
  Duffy}{2021}]{Koratpallikar2021539}
\begin{bchapter}
\bauthor{\bsnm{Koratpallikar}, \binits{P.}},
\bauthor{\bsnm{Duffy}, \binits{V.G.}}:
\bctitle{Cross-cultural design in consumer vehicles to improve safety: A
  systematic literature review}.
In: \bbtitle{Lecture Notes in Computer Science},
vol. \bseriesno{13094 LNCS},
pp. \bfpage{539}--\blpage{553}
(\byear{2021}).
\doiurl{10.1007/978-3-030-90238-4_38}
\end{bchapter}
\endbibitem

\bibitem[\protect\citeauthoryear{Heimg\"{a}rtner}{2021}]{Heimg2021}
\begin{bchapter}
\bauthor{\bsnm{Heimg\"{a}rtner}, \binits{R.}}:
\bctitle{Towards a generic framework for intercultural user interface design to
  evoke positive cross-cultural {UX}}.
In: \bbtitle{Proc. 2021 Int. Conf. on Human-Computer Interaction, HCII'21}.
\bpublisher{Springer},
\blocation{Berlin, Heidelberg}
(\byear{2021}).
\doiurl{10.1007/978-3-030-77431-8_24}
\end{bchapter}
\endbibitem

\bibitem[\protect\citeauthoryear{Miraz et~al.}{2021}]{Miraz2021199}
\begin{barticle}
\bauthor{\bsnm{Miraz}, \binits{M.H.}},
\bauthor{\bsnm{Excell}, \binits{P.S.}},
\bauthor{\bsnm{Ali}, \binits{M.}}:
\batitle{Culturally inclusive adaptive user interface (ciaui) framework:
  Exploration of plasticity of user interface design}.
\bjtitle{Int. J. Inf. Technol. Decis. Mak.}
\bvolume{20}(\bissue{1}),
\bfpage{199}--\blpage{224}
(\byear{2021})
\doiurl{10.1142/S0219622020500455}
\end{barticle}
\endbibitem

\bibitem[\protect\citeauthoryear{Miraz et~al.}{2022}]{Miraz2022}
\begin{botherref}
\oauthor{\bsnm{Miraz}, \binits{M.H.}},
\oauthor{\bsnm{Ali}, \binits{M.}},
\oauthor{\bsnm{Excell}, \binits{P.S.}}:
Cross-cultural usability evaluation of ai-based adaptive user interface for
  mobile applications.
Acta Sci. Technol.
\textbf{44}
(2022)
\doiurl{10.4025/actascitechnol.v44i1.61112}
\end{botherref}
\endbibitem

\bibitem[\protect\citeauthoryear{Singh and Matsuo}{2004}]{SINGH2004864}
\begin{barticle}
\bauthor{\bsnm{Singh}, \binits{N.}},
\bauthor{\bsnm{Matsuo}, \binits{H.}}:
\batitle{Measuring cultural adaptation on the web: a content analytic study of
  u.s. and japanese web sites}.
\bjtitle{J. Bus. Res.}
\bvolume{57}(\bissue{8}),
\bfpage{864}--\blpage{872}
(\byear{2004})
\doiurl{10.1016/S0148-2963(02)00482-4}
\end{barticle}
\endbibitem

\bibitem[\protect\citeauthoryear{Dotan and Zaphiris}{2010}]{Dotan2010284}
\begin{barticle}
\bauthor{\bsnm{Dotan}, \binits{A.}},
\bauthor{\bsnm{Zaphiris}, \binits{P.}}:
\batitle{A cross-cultural analysis of flickr users from peru, israel, iran,
  taiwan and the uk}.
\bjtitle{Int. J. Web Based Communities}
\bvolume{6}(\bissue{3}),
\bfpage{284}--\blpage{302}
(\byear{2010})
\doiurl{10.1504/IJWBC.2010.033753}
\end{barticle}
\endbibitem

\bibitem[\protect\citeauthoryear{Alexander et~al.}{2017}]{Alexander2017}
\begin{bchapter}
\bauthor{\bsnm{Alexander}, \binits{R.}},
\bauthor{\bsnm{Murray}, \binits{D.}},
\bauthor{\bsnm{Thompson}, \binits{N.}}:
\bctitle{Cross-cultural web design guidelines}.
In: \bbtitle{Proc. 14th Web for All Conf., W4A 2017},
\bconflocation{Perth, Australia}
(\byear{2017}).
\doiurl{10.1145/3058555.3058574}
\end{bchapter}
\endbibitem

\bibitem[\protect\citeauthoryear{McMullen}{2016}]{McMullen201619}
\begin{barticle}
\bauthor{\bsnm{McMullen}, \binits{M.}}:
\batitle{Intercultural design competence: A guide for graphic designers working
  across cultural boundaries}.
\bjtitle{Int. J. Vis. Des.}
\bvolume{10}(\bissue{3}),
\bfpage{19}--\blpage{30}
(\byear{2016})
\doiurl{10.18848/2325-1581/cgp/v10i03/19-30}
\end{barticle}
\endbibitem

\bibitem[\protect\citeauthoryear{Li et~al.}{2022}]{Li2022105}
\begin{bchapter}
\bauthor{\bsnm{Li}, \binits{Y.}},
\bauthor{\bsnm{Karreman}, \binits{J.}},
\bauthor{\bsnm{Jong}, \binits{M.D.}}:
\bctitle{Cultural differences in web design on chinese and western websites: A
  literature review}.
In: \bbtitle{Proc. 2022 IEEE Int. Conf. Professional Communication,
  ProComm'22},
vol. \bseriesno{2022}.
\bconflocation{Limerick, Ireland},
pp. \bfpage{105}--\blpage{111}
(\byear{2022}).
\doiurl{10.1109/ProComm53155.2022.00023}
\end{bchapter}
\endbibitem

\bibitem[\protect\citeauthoryear{Bartneck et~al.}{2005}]{Bartneck20051}
\begin{bchapter}
\bauthor{\bsnm{Bartneck}, \binits{C.}},
\bauthor{\bsnm{Nomura}, \binits{T.}},
\bauthor{\bsnm{Kanda}, \binits{T.}},
\bauthor{\bsnm{Suzuki}, \binits{T.}},
\bauthor{\bsnm{Kato}, \binits{K.}}:
\bctitle{Cultural differences in attitudes towards robots}.
In: \bbtitle{Proc. 2005 Symp. on Robot Companions: Hard Problems and Open
  Challenges in Robot-Human Interaction, AISB'05},
\bconflocation{Hatfield, UK},
pp. \bfpage{1}--\blpage{4}
(\byear{2005}).
\doiurl{10.13140/RG.2.2.22507.34085}
\end{bchapter}
\endbibitem

\bibitem[\protect\citeauthoryear{Shibata et~al.}{2009}]{Shibata2009443}
\begin{barticle}
\bauthor{\bsnm{Shibata}, \binits{T.}},
\bauthor{\bsnm{Wada}, \binits{K.}},
\bauthor{\bsnm{Ikeda}, \binits{Y.}},
\bauthor{\bsnm{Sabanovic}, \binits{S.}}:
\batitle{Cross-cultural studies on subjective evaluation of a seal robot}.
\bjtitle{Adv. Robot.}
\bvolume{23}(\bissue{4}),
\bfpage{443}--\blpage{458}
(\byear{2009})
\doiurl{10.1163/156855309X408826}
\end{barticle}
\endbibitem

\bibitem[\protect\citeauthoryear{Rau et~al.}{2010}]{Rau2010175}
\begin{barticle}
\bauthor{\bsnm{Rau}, \binits{P.L.P.}},
\bauthor{\bsnm{Li}, \binits{Y.}},
\bauthor{\bsnm{Li}, \binits{D.}}:
\batitle{A cross-cultural study: Effect of robot appearance and task}.
\bjtitle{Int. J. Soc. Robot.}
\bvolume{2}(\bissue{2}),
\bfpage{175}--\blpage{186}
(\byear{2010})
\doiurl{10.1007/s12369-010-0056-9}
\end{barticle}
\endbibitem

\bibitem[\protect\citeauthoryear{Mavridis et~al.}{2012}]{Mavridis2012517}
\begin{barticle}
\bauthor{\bsnm{Mavridis}, \binits{N.}},
\bauthor{\bsnm{Katsaiti}, \binits{M.-S.}},
\bauthor{\bsnm{Naef}, \binits{S.}},
\bauthor{\bsnm{Falasi}, \binits{A.}},
\bauthor{\bsnm{Nuaimi}, \binits{A.}},
\bauthor{\bsnm{Araifi}, \binits{H.}},
\bauthor{\bsnm{Kitbi}, \binits{A.}}:
\batitle{Opinions and attitudes toward humanoid robots in the middle east}.
\bjtitle{AI Soc.}
\bvolume{27}(\bissue{4}),
\bfpage{517}--\blpage{534}
(\byear{2012})
\doiurl{10.1007/s00146-011-0370-2}
\end{barticle}
\endbibitem

\bibitem[\protect\citeauthoryear{Nomura et~al.}{2015}]{Nomura2015}
\begin{bchapter}
\bauthor{\bsnm{Nomura}, \binits{T.}},
\bauthor{\bsnm{Syrdal}, \binits{D.S.}},
\bauthor{\bsnm{Dautenhahn}, \binits{K.}}:
\bctitle{Differences on social acceptance of humanoid robots between japan and
  the uk}.
In: \bbtitle{Proc. 4th Int. Symp. on New Frontiers in Human-Robot Interaction,
  AISB'15},
\bconflocation{Canterbury, UK}
(\byear{2015})
\end{bchapter}
\endbibitem

\bibitem[\protect\citeauthoryear{Haring et~al.}{2014}]{Haring2014166}
\begin{bchapter}
\bauthor{\bsnm{Haring}, \binits{K.S.}},
\bauthor{\bsnm{Silvera-Tawil}, \binits{D.}},
\bauthor{\bsnm{Matsumoto}, \binits{Y.}},
\bauthor{\bsnm{Velonaki}, \binits{M.}},
\bauthor{\bsnm{Watanabe}, \binits{K.}}:
\bctitle{Perception of an android robot in japan and australia: A
  cross-cultural comparison}.
In: \bbtitle{Lecture Notes in Computer Science},
vol. \bseriesno{8755 LNCS},
pp. \bfpage{166}--\blpage{175}
(\byear{2014}).
\doiurl{10.1007/978-3-319-11973-1_17}
\end{bchapter}
\endbibitem

\bibitem[\protect\citeauthoryear{Trovato et~al.}{2018}]{Trovato2018}
\begin{botherref}
\oauthor{\bsnm{Trovato}, \binits{G.}},
\oauthor{\bsnm{Lucho}, \binits{C.}},
\oauthor{\bsnm{Paredes}, \binits{R.}}:
She's electric-the influence of body proportions on perceived gender of robots
  across cultures.
Robotics
\textbf{7}(3)
(2018)
\doiurl{10.3390/robotics7030050}
\end{botherref}
\endbibitem

\bibitem[\protect\citeauthoryear{Berque et~al.}{2022}]{Berque2022391}
\begin{bchapter}
\bauthor{\bsnm{Berque}, \binits{D.}},
\bauthor{\bsnm{Chiba}, \binits{H.}},
\bauthor{\bsnm{Laohakangvalvit}, \binits{T.}},
\bauthor{\bsnm{Ohkura}, \binits{M.}},
\bauthor{\bsnm{Sripian}, \binits{P.}},
\bauthor{\bsnm{Sugaya}, \binits{M.}},
\bauthor{\bsnm{Guinee}, \binits{L.}},
\bauthor{\bsnm{Imura}, \binits{S.}},
\bauthor{\bsnm{Jadram}, \binits{N.}},
\bauthor{\bsnm{Martinez}, \binits{R.}},
\bauthor{\bsnm{Ng}, \binits{S.F.}},
\bauthor{\bsnm{Schwipps}, \binits{H.}},
\bauthor{\bsnm{Ohtsuka}, \binits{S.}},
\bauthor{\bsnm{Todd}, \binits{G.}}:
\bctitle{Cross-cultural design and evaluation of student companion robots with
  varied kawaii (cute) attributes}.
In: \bbtitle{Lecture Notes in Computer Science},
vol. \bseriesno{13302 LNCS},
pp. \bfpage{391}--\blpage{409}
(\byear{2022}).
\doiurl{10.1007/978-3-031-05311-5_27}
\end{bchapter}
\endbibitem

\bibitem[\protect\citeauthoryear{Dang and Liu}{2023}]{Dang2023}
\begin{botherref}
\oauthor{\bsnm{Dang}, \binits{J.}},
\oauthor{\bsnm{Liu}, \binits{L.}}:
Do lonely people seek robot companionship? a comparative examination of the
  loneliness–robot anthropomorphism link in the united states and china.
Comput. Hum. Behav.
\textbf{141}
(2023)
\doiurl{10.1016/j.chb.2022.107637}
\end{botherref}
\endbibitem

\bibitem[\protect\citeauthoryear{Bernsteiner et~al.}{2022}]{Bernsteiner2022517}
\begin{barticle}
\bauthor{\bsnm{Bernsteiner}, \binits{A.}},
\bauthor{\bsnm{Pollmann}, \binits{K.}},
\bauthor{\bsnm{Neuhold}, \binits{L.}}:
\batitle{Comparative study on the impact of cultural background on the
  perception of different types of social robots}.
\bjtitle{Commun. Comput. Inf. Sci.}
\bvolume{1655 CCIS},
\bfpage{517}--\blpage{522}
(\byear{2022})
\doiurl{10.1007/978-3-031-19682-9_65}
\end{barticle}
\endbibitem

\bibitem[\protect\citeauthoryear{Wang et~al.}{2010}]{Wang2010359}
\begin{bchapter}
\bauthor{\bsnm{Wang}, \binits{L.}},
\bauthor{\bsnm{Rau}, \binits{P.-L.P.}},
\bauthor{\bsnm{Evers}, \binits{V.}},
\bauthor{\bsnm{Robinson}, \binits{B.K.}},
\bauthor{\bsnm{Hinds}, \binits{P.}}:
\bctitle{When in rome: The role of culture \& context in adherence to robot
  recommendations}.
In: \bbtitle{Proc. 5th ACM/IEEE Int. Conf. on Human-Robot Interaction, HRI'10},
\bconflocation{Osaka, Japan},
pp. \bfpage{359}--\blpage{366}
(\byear{2010}).
\doiurl{10.1145/1734454.1734578}
\end{bchapter}
\endbibitem

\bibitem[\protect\citeauthoryear{Shidujaman et~al.}{2020}]{Shidujaman2020}
\begin{bchapter}
\bauthor{\bsnm{Shidujaman}, \binits{M.}},
\bauthor{\bsnm{Mi}, \binits{H.}},
\bauthor{\bsnm{Jamal}, \binits{L.}}:
\bctitle{"I Trust You More": A Behavioral Greeting Gesture Study on Social
  Robots for Recommendation Tasks}.
In: \bbtitle{Proc. 2020 Int. Conf. on Image Processing and Robotics, ICIPRoB
  2020},
\blocation{Negombo, Sri Lanka}
(\byear{2020}).
\doiurl{10.1109/ICIP48927.2020.9367364}
\end{bchapter}
\endbibitem

\bibitem[\protect\citeauthoryear{Bliss et~al.}{2020}]{Bliss2020493}
\begin{bchapter}
\bauthor{\bsnm{Bliss}, \binits{J.P.}},
\bauthor{\bsnm{Gao}, \binits{Q.}},
\bauthor{\bsnm{Hu}, \binits{X.}},
\bauthor{\bsnm{Itoh}, \binits{M.}},
\bauthor{\bsnm{Karpinsky-Mosely}, \binits{N.}},
\bauthor{\bsnm{Long}, \binits{S.K.}},
\bauthor{\bsnm{Papelis}, \binits{Y.}},
\bauthor{\bsnm{Yamani}, \binits{Y.}}:
\bctitle{Cross-cultural trust of robot peacekeepers as a function of dialog,
  appearance, responsibilities, and onboard weapons}.
\bbtitle{Trust in Human-Robot Interaction},
pp. \bfpage{493}--\blpage{513}.
\bpublisher{Elsevier}, \blocation{???}
(\byear{2020}).
\doiurl{10.1016/B978-0-12-819472-0.00021-6}
\end{bchapter}
\endbibitem

\bibitem[\protect\citeauthoryear{Andrist et~al.}{2015}]{Andrist2015157}
\begin{bchapter}
\bauthor{\bsnm{Andrist}, \binits{S.}},
\bauthor{\bsnm{Ziadee}, \binits{M.}},
\bauthor{\bsnm{Boukaram}, \binits{H.}},
\bauthor{\bsnm{Mutlu}, \binits{B.}},
\bauthor{\bsnm{Sakr}, \binits{M.}}:
\bctitle{Effects of culture on the credibility of robot speech: A comparison
  between english and arabic}.
In: \bbtitle{Proc. 2015 ACM/IEEE Int. Conf. on Human-Robot Interaction,
  HRI'14},
vol. \bseriesno{2015-March}.
\bconflocation{Portland, Ore, USA},
pp. \bfpage{157}--\blpage{164}
(\byear{2015}).
\doiurl{10.1145/2696454.2696464}
\end{bchapter}
\endbibitem

\bibitem[\protect\citeauthoryear{Rudovic et~al.}{2017}]{Rudovic2017}
\begin{botherref}
\oauthor{\bsnm{Rudovic}, \binits{O.O.}},
\oauthor{\bsnm{Lee}, \binits{J.}},
\oauthor{\bsnm{Mascarell-Maricic}, \binits{L.}},
\oauthor{\bsnm{Schuller}, \binits{B.W.}},
\oauthor{\bsnm{Picard}, \binits{R.W.}}:
Measuring engagement in robotassisted autism therapy: A crosscultural study.
Front. Robot. AI
\textbf{4}
(2017)
\doiurl{10.3389/frobt.2017.00036}
\end{botherref}
\endbibitem

\bibitem[\protect\citeauthoryear{Makenova et~al.}{2018}]{Makenova2018185}
\begin{bchapter}
\bauthor{\bsnm{Makenova}, \binits{R.}},
\bauthor{\bsnm{Karsybayeva}, \binits{R.}},
\bauthor{\bsnm{Sandygulova}, \binits{A.}}:
\bctitle{Exploring cross-cultural differences in persuasive robotics}.
In: \bbtitle{Proc. 2018 ACM/IEEE Int. Conf. on Human-Robot Interaction,
  HRI'18},
\bconflocation{Chicago, IL, USA},
pp. \bfpage{185}--\blpage{186}
(\byear{2018}).
\doiurl{10.1145/3173386.3177079}
\end{bchapter}
\endbibitem

\bibitem[\protect\citeauthoryear{Knight et~al.}{2009}]{Knight200917}
\begin{barticle}
\bauthor{\bsnm{Knight}, \binits{E.}},
\bauthor{\bsnm{Gunawardena}, \binits{C.N.}},
\bauthor{\bsnm{Aydin}, \binits{C.H.}}:
\batitle{Cultural interpretations of the visual meaning of icons and images
  used in north american web design}.
\bjtitle{Educ. Media Int.}
\bvolume{46}(\bissue{1}),
\bfpage{17}--\blpage{35}
(\byear{2009})
\doiurl{10.1080/09523980902781279}
\end{barticle}
\endbibitem

\bibitem[\protect\citeauthoryear{McMullen}{2019}]{McMullen20191}
\begin{barticle}
\bauthor{\bsnm{McMullen}, \binits{M.}}:
\batitle{Cross-cultural design: Identifying cultural markers of printed graphic
  design from germany and south korea}.
\bjtitle{Int. J. Vis. Des.}
\bvolume{13}(\bissue{2}),
\bfpage{1}--\blpage{15}
(\byear{2019})
\doiurl{10.18848/2325-1581/CGP/v13i02/1-15}
\end{barticle}
\endbibitem

\bibitem[\protect\citeauthoryear{Quan et~al.}{2018}]{Quan2018}
\begin{botherref}
\oauthor{\bsnm{Quan}, \binits{H.}},
\oauthor{\bsnm{Li}, \binits{S.}},
\oauthor{\bsnm{Hu}, \binits{J.}}:
Product innovation design based on deep learning and kansei engineering.
Appl. Sci.
\textbf{8}(12)
(2018)
\doiurl{10.3390/app8122397}
\end{botherref}
\endbibitem

\bibitem[\protect\citeauthoryear{Yanlong}{2021}]{Yanlong2021721}
\begin{bchapter}
\bauthor{\bsnm{Yanlong}, \binits{G.}}:
\bctitle{Application of neural style transfer and image recognition in the
  computer aided design of creative products adopting computational vision}.
In: \bbtitle{Proc. 2021 IEEE Int. Conf. on Emergency Science and Information
  Technology, ICESIT 2021},
\bconflocation{Chongqing, China},
pp. \bfpage{721}--\blpage{724}
(\byear{2021}).
\doiurl{10.1109/ICESIT53460.2021.9696993}
\end{bchapter}
\endbibitem

\bibitem[\protect\citeauthoryear{Xuelin et~al.}{2021}]{Xuelin2021339}
\begin{bchapter}
\bauthor{\bsnm{Xuelin}, \binits{Q.}},
\bauthor{\bsnm{Jue}, \binits{H.}},
\bauthor{\bsnm{Ying}, \binits{S.}},
\bauthor{\bsnm{Zheng}, \binits{L.}}:
\bctitle{Digital style design of {Nanjing} brocade based on deep learning}.
In: \bbtitle{Proc. 2021 Int. Conf. on Culture-Oriented Science and Technology,
  ICCST 2021},
\bconflocation{Beijing, China},
pp. \bfpage{339}--\blpage{342}
(\byear{2021}).
\doiurl{10.1109/ICCST53801.2021.00077}
\end{bchapter}
\endbibitem

\bibitem[\protect\citeauthoryear{Joseph et~al.}{2021}]{Joseph2021}
\begin{bchapter}
\bauthor{\bsnm{Joseph}, \binits{M.}},
\bauthor{\bsnm{Richard}, \binits{J.}},
\bauthor{\bsnm{Halim}, \binits{C.S.}},
\bauthor{\bsnm{Faadhilah}, \binits{R.}},
\bauthor{\bsnm{Qomariyah}, \binits{N.N.}}:
\bctitle{Recreating traditional indonesian batik with neural style transfer in
  ai artistry}.
In: \bbtitle{Proc. 8th Int. Conf. on ICT for Smart Society: Digital Twin for
  Smart Society, ICISS 2021},
\bconflocation{Bandung, Indonesia}
(\byear{2021}).
\doiurl{10.1109/ICISS53185.2021.9533197}
\end{bchapter}
\endbibitem

\bibitem[\protect\citeauthoryear{Liu et~al.}{2021}]{Liu2021}
\begin{botherref}
\oauthor{\bsnm{Liu}, \binits{S.}},
\oauthor{\bsnm{Bo}, \binits{Y.}},
\oauthor{\bsnm{Huang}, \binits{L.}}:
Application of image style transfer technology in interior decoration design
  based on ecological environment.
J. Sens.
\textbf{2021}
(2021)
\doiurl{10.1155/2021/9699110}
\end{botherref}
\endbibitem

\bibitem[\protect\citeauthoryear{Wu et~al.}{2021}]{Wu2021341}
\begin{barticle}
\bauthor{\bsnm{Wu}, \binits{Q.}},
\bauthor{\bsnm{Zhu}, \binits{B.}},
\bauthor{\bsnm{Yong}, \binits{B.}},
\bauthor{\bsnm{Wei}, \binits{Y.}},
\bauthor{\bsnm{Jiang}, \binits{X.}},
\bauthor{\bsnm{Zhou}, \binits{R.}},
\bauthor{\bsnm{Zhou}, \binits{Q.}}:
\batitle{Clothgan: generation of fashionable dunhuang clothes using generative
  adversarial networks}.
\bjtitle{Conn. Sci.}
\bvolume{33}(\bissue{2}),
\bfpage{341}--\blpage{358}
(\byear{2021})
\doiurl{10.1080/09540091.2020.1822780}
\end{barticle}
\endbibitem

\bibitem[\protect\citeauthoryear{Zhang and Romainoor}{2023}]{Zhang2023}
\begin{botherref}
\oauthor{\bsnm{Zhang}, \binits{B.}},
\oauthor{\bsnm{Romainoor}, \binits{N.H.}}:
Research on artificial intelligence in new year prints: The application of the
  generated pop art style images on cultural and creative products.
Appl. Sci.
\textbf{13}(2)
(2023)
\doiurl{10.3390/app13021082}
\end{botherref}
\endbibitem

\bibitem[\protect\citeauthoryear{Fu et~al.}{2022}]{Fu2022670}
\begin{bchapter}
\bauthor{\bsnm{Fu}, \binits{R.}},
\bauthor{\bsnm{Wang}, \binits{Y.}},
\bauthor{\bsnm{Lin}, \binits{J.}},
\bauthor{\bsnm{Fan}, \binits{R.}},
\bauthor{\bsnm{Zhao}, \binits{H.}}:
\bctitle{Shanghai-style realistic style watercolor painting style transfer by
  using rsim evaluation}.
In: \bbtitle{Lecture Notes in Electrical Engineering},
vol. \bseriesno{801 LNEE},
pp. \bfpage{670}--\blpage{682}
(\byear{2022}).
\doiurl{10.1007/978-981-16-6372-7_72}
\end{bchapter}
\endbibitem

\bibitem[\protect\citeauthoryear{Zhou et~al.}{2022}]{Zhou2022445}
\begin{barticle}
\bauthor{\bsnm{Zhou}, \binits{L.}},
\bauthor{\bsnm{Sun}, \binits{X.}},
\bauthor{\bsnm{Mu}, \binits{G.}},
\bauthor{\bsnm{Wu}, \binits{J.}},
\bauthor{\bsnm{Zhou}, \binits{J.}},
\bauthor{\bsnm{Wu}, \binits{Q.}},
\bauthor{\bsnm{Zhang}, \binits{Y.}},
\bauthor{\bsnm{Xi}, \binits{Y.}},
\bauthor{\bsnm{Gunes}, \binits{N.D.}},
\bauthor{\bsnm{Song}, \binits{S.}}:
\batitle{A tool to facilitate the cross-cultural design process using deep
  learning}.
\bjtitle{IEEE Trans. Hum.-Mach. Syst.}
\bvolume{52}(\bissue{3}),
\bfpage{445}--\blpage{457}
(\byear{2022})
\end{barticle}
\endbibitem

\bibitem[\protect\citeauthoryear{Kaji and Kida}{2019}]{Kaji2019}
\begin{barticle}
\bauthor{\bsnm{Kaji}, \binits{S.}},
\bauthor{\bsnm{Kida}, \binits{S.}}:
\batitle{Overview of image-to-image translation by use of deep neural networks:
  denoising, super-resolution, modality conversion, and reconstruction in
  medical imaging}.
\bjtitle{Radiol. Phys. and Technol.}
\bvolume{12}(\bissue{3}),
\bfpage{235}--\blpage{248}
(\byear{2019})
\doiurl{10.1007/s12194-019-00520-y}
\end{barticle}
\endbibitem

\bibitem[\protect\citeauthoryear{Pang et~al.}{2021}]{pang2021imagetoimage}
\begin{barticle}
\bauthor{\bsnm{Pang}, \binits{Y.}},
\bauthor{\bsnm{Lin}, \binits{J.}},
\bauthor{\bsnm{Qin}, \binits{T.}},
\bauthor{\bsnm{Chen}, \binits{Z.}}:
\batitle{Image-to-image translation: Methods and applications}.
\bjtitle{eprint arXiv:2101.08629}
(\byear{2021})
\doiurl{10.48550/arXiv.2101.08629}
\end{barticle}
\endbibitem

\bibitem[\protect\citeauthoryear{Radford
  et~al.}{2016}]{radford2015unsupervised}
\begin{bchapter}
\bauthor{\bsnm{Radford}, \binits{A.}},
\bauthor{\bsnm{Metz}, \binits{L.}},
\bauthor{\bsnm{Chintala}, \binits{S.}}:
\bctitle{Unsupervised representation learning with deep convolutional
  generative adversarial networks}.
In: \bbtitle{{Proc. 2016 Int. Conf. on Learning Representations, ICLR}'16},
\bconflocation{San Juan, Puerto Rico}
(\byear{2016}).
\doiurl{10.48550/arXiv.1511.06434}
\end{bchapter}
\endbibitem

\bibitem[\protect\citeauthoryear{Huang et~al.}{2021}]{Huang2021}
\begin{barticle}
\bauthor{\bsnm{Huang}, \binits{Z.}},
\bauthor{\bsnm{Chen}, \binits{S.}},
\bauthor{\bsnm{Zhang}, \binits{J.}},
\bauthor{\bsnm{Shan}, \binits{H.}}:
\batitle{{PFA-GAN}: Progressive face aging with generative adversarial
  network}.
\bjtitle{IEEE Trans. Inf. Forensics Secur.}
\bvolume{16},
\bfpage{2031}--\blpage{2045}
(\byear{2021})
\doiurl{10.1109/TIFS.2020.3047753}
\end{barticle}
\endbibitem

\bibitem[\protect\citeauthoryear{Eitz et~al.}{2012}]{Eitz2012}
\begin{barticle}
\bauthor{\bsnm{Eitz}, \binits{M.}},
\bauthor{\bsnm{Hays}, \binits{J.}},
\bauthor{\bsnm{Alexa}, \binits{M.}}:
\batitle{How do humans sketch objects?}
\bjtitle{ACM Trans. Graph.}
\bvolume{31},
\bfpage{1}--\blpage{10}
(\byear{2012})
\doiurl{10.1145/2185520.2185540}
\end{barticle}
\endbibitem

\bibitem[\protect\citeauthoryear{Cordts et~al.}{2016}]{Cordts2016}
\begin{bchapter}
\bauthor{\bsnm{Cordts}, \binits{M.}},
\bauthor{\bsnm{Omran}, \binits{M.}},
\bauthor{\bsnm{Ramos}, \binits{S.}},
\bauthor{\bsnm{Rehfeld}, \binits{T.}},
\bauthor{\bsnm{Enzweiler}, \binits{M.}},
\bauthor{\bsnm{Benenson}, \binits{R.}},
\bauthor{\bsnm{Franke}, \binits{U.}},
\bauthor{\bsnm{Roth}, \binits{S.}},
\bauthor{\bsnm{Schiele}, \binits{B.}}:
\bctitle{The cityscapes dataset for semantic urban scene understanding}.
In: \bbtitle{Proc. 2016 Conf. on Computer Vision and Pattern Recognition,
  CVPR'16},
\bconflocation{Las Vegas, NV, USA},
pp. \bfpage{3213}--\blpage{3223}
(\byear{2016}).
\doiurl{10.1109/CVPR.2016.350}
\end{bchapter}
\endbibitem

\bibitem[\protect\citeauthoryear{Li et~al.}{2019}]{Haofeng2019}
\begin{barticle}
\bauthor{\bsnm{Li}, \binits{H.}},
\bauthor{\bsnm{Li}, \binits{G.}},
\bauthor{\bsnm{Lin}, \binits{L.}},
\bauthor{\bsnm{Yu}, \binits{H.}},
\bauthor{\bsnm{Yu}, \binits{Y.}}:
\batitle{Context-aware semantic inpainting}.
\bjtitle{IEEE Trans. Cybern.}
\bvolume{49}(\bissue{12}),
\bfpage{4398}--\blpage{4411}
(\byear{2019})
\doiurl{10.1109/TCYB.2018.2865036}
\end{barticle}
\endbibitem

\bibitem[\protect\citeauthoryear{Mirza and
  Osindero}{2014}]{mirza2014conditional}
\begin{barticle}
\bauthor{\bsnm{Mirza}, \binits{M.}},
\bauthor{\bsnm{Osindero}, \binits{S.}}:
\batitle{Conditional generative adversarial nets}.
\bjtitle{eprint arXiv:1411.1784}
(\byear{2014})
\doiurl{10.48550/arXiv.1411.1}
\end{barticle}
\endbibitem

\bibitem[\protect\citeauthoryear{Yi et~al.}{2017}]{Zili2017}
\begin{bchapter}
\bauthor{\bsnm{Yi}, \binits{Z.}},
\bauthor{\bsnm{Zhang}, \binits{H.}},
\bauthor{\bsnm{Tan}, \binits{P.}},
\bauthor{\bsnm{Gong}, \binits{M.}}:
\bctitle{{DualGAN}: Unsupervised dual learning for image-to-image translation}.
In: \bbtitle{Proc. 2017 Int. Conf. on Computer Vision, {ICCV}'17},
\bconflocation{Venice, Italy}
(\byear{2017}).
\doiurl{10.1109/ICCV.2017.310}
\end{bchapter}
\endbibitem

\bibitem[\protect\citeauthoryear{Chen et~al.}{2018}]{Chen2018}
\begin{barticle}
\bauthor{\bsnm{Chen}, \binits{X.}},
\bauthor{\bsnm{Xu}, \binits{C.}},
\bauthor{\bsnm{Yang}, \binits{X.}},
\bauthor{\bsnm{Tao}, \binits{D.}}:
\batitle{Attention-{GAN} for object transfiguration in wild images}.
\bjtitle{Lecture Notes In Computer Science}
\bvolume{11206 LNCS},
\bfpage{167}--\blpage{184}
(\byear{2018})
\doiurl{10.1007/978-3-030-01216-8_11}
\end{barticle}
\endbibitem

\bibitem[\protect\citeauthoryear{Tang et~al.}{2021}]{Tang2021}
\begin{barticle}
\bauthor{\bsnm{Tang}, \binits{H.}},
\bauthor{\bsnm{Liu}, \binits{H.}},
\bauthor{\bsnm{Xu}, \binits{D.}},
\bauthor{\bsnm{Torr}, \binits{P.H.S.}},
\bauthor{\bsnm{Sebe}, \binits{N.}}:
\batitle{{AttentionGAN}: Unpaired image-to-image translation using
  attention-guided generative adversarial networks}.
\bjtitle{IEEE Trans. Neural Netw. Learn. Syst.}
(\byear{2021})
\doiurl{10.1109/TNNLS.2021.3105725}
\end{barticle}
\endbibitem

\bibitem[\protect\citeauthoryear{Liu and Tuzel}{2016}]{Ming2016}
\begin{bchapter}
\bauthor{\bsnm{Liu}, \binits{M.}},
\bauthor{\bsnm{Tuzel}, \binits{O.}}:
\bctitle{Coupled generative adversarial networks}.
In: \bbtitle{Proc. 2016 Conf. on Neural Information Processing Systems,
  NIPS'16},
\bconflocation{Barcelona, Spain}
(\byear{2016}).
\doiurl{10.48550/arXiv.1606.07536}
\end{bchapter}
\endbibitem

\bibitem[\protect\citeauthoryear{Larsen et~al.}{2016}]{VAeGAN}
\begin{bchapter}
\bauthor{\bsnm{Larsen}, \binits{A.B.L.}},
\bauthor{\bsnm{S\o{}nderby}, \binits{S.K.}},
\bauthor{\bsnm{Larochelle}, \binits{H.}},
\bauthor{\bsnm{Winther}, \binits{O.}}:
\bctitle{Autoencoding beyond pixels using a learned similarity metric}.
In: \bbtitle{Proc. 2016 Int. Conf. on Machine Learning, ICML'16},
\bconflocation{New York, NY, USA},
pp. \bfpage{1558}--\blpage{1566}
(\byear{2016}).
\doiurl{10.48550/arXiv.1512.09300}
\end{bchapter}
\endbibitem

\bibitem[\protect\citeauthoryear{Borji}{2019}]{Proecons2018}
\begin{barticle}
\bauthor{\bsnm{Borji}, \binits{A.}}:
\batitle{Pros and cons of {GAN} evaluation measures}.
\bjtitle{Comput. Vis. and Image Underst.}
\bvolume{179},
\bfpage{41}--\blpage{65}
(\byear{2019})
\doiurl{10.1016/j.cviu.2018.10.009}
\end{barticle}
\endbibitem

\bibitem[\protect\citeauthoryear{Salimans et~al.}{2016}]{Tim2016}
\begin{bchapter}
\bauthor{\bsnm{Salimans}, \binits{T.}},
\bauthor{\bsnm{Goodfellow}, \binits{I.J.}},
\bauthor{\bsnm{Zaremba}, \binits{W.}},
\bauthor{\bsnm{Cheung}, \binits{V.}},
\bauthor{\bsnm{Radford}, \binits{A.}},
\bauthor{\bsnm{Chen}, \binits{X.}}:
\bctitle{Improved techniques for training {GANs}}.
In: \bbtitle{Proc. 2016 Conf. on Neural Information Processing Systems,
  NIPS'16},
\bconflocation{Barcelona, Spain}
(\byear{2016}).
\doiurl{10.48550/arXiv.1606.03498}
\end{bchapter}
\endbibitem

\bibitem[\protect\citeauthoryear{Heusel et~al.}{2017}]{FID2017}
\begin{bchapter}
\bauthor{\bsnm{Heusel}, \binits{M.}},
\bauthor{\bsnm{Ramsauer}, \binits{H.}},
\bauthor{\bsnm{Unterthiner}, \binits{T.}},
\bauthor{\bsnm{Nessler}, \binits{B.}},
\bauthor{\bsnm{Klambauer}, \binits{G.}},
\bauthor{\bsnm{Hochreiter}, \binits{S.}}:
\bctitle{{GANs} trained by a two time-scale update rule converge to a {Nash}
  equilibrium}.
In: \bbtitle{Proc. 2017 Conf. on Neural Information Processing Systems,
  NIPS'17},
\bconflocation{Long Beach, CA, USA}
(\byear{2017}).
\doiurl{10.48550/arXiv.1706.08500}
\end{bchapter}
\endbibitem

\bibitem[\protect\citeauthoryear{Szegedy et~al.}{2016}]{Szegedy20162818}
\begin{bchapter}
\bauthor{\bsnm{Szegedy}, \binits{C.}},
\bauthor{\bsnm{Vanhoucke}, \binits{V.}},
\bauthor{\bsnm{Ioffe}, \binits{S.}},
\bauthor{\bsnm{Shlens}, \binits{J.}},
\bauthor{\bsnm{Wojna}, \binits{Z.}}:
\bctitle{Rethinking the inception architecture for computer vision}.
In: \bbtitle{Proc. 2016 IEEE Conf. on Computer Vision and Pattern Recognition,
  CVPR'16},
\bconflocation{Las Vegas, NV, USA},
pp. \bfpage{2818}--\blpage{2826}
(\byear{2016}).
\doiurl{10.1109/CVPR.2016.308}
\end{bchapter}
\endbibitem

\bibitem[\protect\citeauthoryear{Zhao et~al.}{2020}]{Zhao2020}
\begin{bchapter}
\bauthor{\bsnm{Zhao}, \binits{Y.}},
\bauthor{\bsnm{Wu}, \binits{R.}},
\bauthor{\bsnm{Dong}, \binits{H.}}:
\bctitle{Unpaired image-to-image translation using adversarial consistency
  loss}.
In: \bbtitle{Proc. 2010 Europ. Conf. on Computer Vision, ECCV 2020},
\bconflocation{Glasgow, UK},
pp. \bfpage{800}--\blpage{815}
(\byear{2020}).
\doiurl{10.1007/978-3-030-58545-7_46}
\end{bchapter}
\endbibitem

\bibitem[\protect\citeauthoryear{Schwaba et~al.}{2018}]{Shwaba17}
\begin{barticle}
\bauthor{\bsnm{Schwaba}, \binits{T.}},
\bauthor{\bsnm{Luhmann}, \binits{M.}},
\bauthor{\bsnm{Denissen}, \binits{J.J.A.}},
\bauthor{\bsnm{Chung}, \binits{J.M.H.}},
\bauthor{\bsnm{Bleidorn}, \binits{W.}}:
\batitle{Openness to experience and culture-openness transactions across the
  lifespan}.
\bjtitle{J. Pers. Soc. Psychol.}
\bvolume{115}(\bissue{1}),
\bfpage{118}--\blpage{136}
(\byear{2018})
\doiurl{10.1037/pspp0000150}
\end{barticle}
\endbibitem

\end{thebibliography}
\end{document}